\documentclass[12pt]{article}
\usepackage{amsmath,amssymb,amsthm,amsfonts,bm,mathrsfs,float,caption,subcaption,setspace,graphicx,comment}
\usepackage{avant,array,geometry,fontenc,inputenc,natbib,hyperref,multirow,enumerate,rotating,booktabs,color,latexsym,times,stmaryrd,xr,authblk}
\usepackage{multicol}
\usepackage{enumitem}
\usepackage{algorithm}
\usepackage{algpseudocode}

\newtheorem{theorem}{Theorem}

\newtheorem{asump}{Assumption}

\DeclareMathOperator*{\argmin}{arg\,min}

\newcommand{\blind}{1}

\NewDocumentCommand{\Rijt}{O{i}O{m}O{t}}{%
    R^{(#1,#2)}_{#3}
}
\NewDocumentCommand{\rijt}{O{i}O{m}O{t}}{%
    r^{(#1, #2)}_{#3}
}
\NewDocumentCommand{\fijt}{O{i}O{m}O{t}}{%
    f^{(#1,#2)}_{#3}
}
\NewDocumentCommand{\deltaijt}{O{i}O{m}O{t}}{%
    \delta^{(#1, #2)}_{#3}
}
\NewDocumentCommand{\xiijt}{O{i}O{m}O{t}}{%
    \xi^{(#1,#2)}_{#3}
}
\NewDocumentCommand{\epsilonijt}{O{i}O{m}O{t}}{%
    \epsilon^{(#1,#2)}_{#3}
}
\NewDocumentCommand{\Aijt}{O{i}O{m}O{t}}{%
    A^{(#1,#2)}_{#3}
}
\NewDocumentCommand{\sijt}{O{i}O{m}O{t}}{%
    s^{(#1,#2)}_{#3}
}
\NewDocumentCommand{\nsijt}{O{i}O{m}O{t}}{%
    s^{(#1,#2) \prime}_{#3}
}
\NewDocumentCommand{\aijt}{O{i}O{m}O{t}}{%
    a^{(#1,#2)}_{#3}
}
\NewDocumentCommand{\Sijt}{O{i}O{m}O{t}}{%
    S^{(#1,#2)}_{#3}
}
\NewDocumentCommand{\TDijt}{O{i}O{m}O{t}}{%
    TD^{(#1,#2)}_{#3}
}

\NewDocumentCommand{\Rit}{O{i}O{t}}{%
    R^{(#1)}_{#2}
}
\NewDocumentCommand{\rit}{O{i}O{t}}{%
    r^{(#1)}_{#2}
}
\NewDocumentCommand{\fit}{O{i}O{t}}{%
    f^{(#1)}_{#2}
}
\NewDocumentCommand{\deltait}{O{i}O{t}}{%
    \delta^{(#1)}_{#2}
}
\NewDocumentCommand{\xiit}{O{i}O{t}}{%
    \xi^{(#1)}_{#2}
}
\NewDocumentCommand{\epsilonit}{O{i}O{t}}{%
    \epsilon^{(#1)}_{#2}
}
\NewDocumentCommand{\Ait}{O{i}O{t}}{%
    A^{(#1)}_{#2}
}
\NewDocumentCommand{\sit}{O{i}O{t}}{%
    s^{(#1)}_{#2}
}
\NewDocumentCommand{\nsit}{O{i}O{t}}{%
    s^{(#1) \prime}_{#2}
}
\NewDocumentCommand{\ait}{O{i}O{t}}{%
    a^{(#1)}_{#2}
}
\NewDocumentCommand{\Sit}{O{i}O{t}}{%
    S^{(#1)}_{#2}
}
\NewDocumentCommand{\TDit}{O{i}O{t}}{%
    TD^{(#1)}_{#2}
}

\newcommand{\bfS}{\mathbf{S}}
\newcommand{\bfA}{\mathbf{A}}
\newcommand{\bfR}{\mathbf{R}}

\newcommand{\bfV}{\mathbf{V}}

\NewDocumentCommand{\Ri}{O{i}}{%
    \mathbf{R}_{#1}
}
\NewDocumentCommand{\ri}{O{i}}{%
    \mathbf{r}_{#1}
}
\NewDocumentCommand{\Si}{O{i}}{%
    \mathbf{S}_{#1}
}
\NewDocumentCommand{\Ai}{O{i}}{%
    \mathbf{A}_{#1}
}
\NewDocumentCommand{\si}{O{i}}{%
    \mathbf{s}_{#1}
}
\NewDocumentCommand{\ai}{O{i}}{%
    \mathbf{a}_{#1}
}
\NewDocumentCommand{\myfi}{O{i}}{%
    \mathbf{f}_{#1}
}
\NewDocumentCommand{\TDi}{O{i}}{%
    \mathbf{TD}_{#1}
}
\NewDocumentCommand{\Yi}{O{i}}{%
    \mathbf{Y}_{#1}
}
\NewDocumentCommand{\Vi}{O{i}}{
    \mathbf{V}_{#1}
}
\NewDocumentCommand{\Di}{O{i}}{
    \mathbf{D}_{#1}
}
\NewDocumentCommand{\Gi}{O{i}}{%
    \mathbf{G}_{#1}
}
\NewDocumentCommand{\Omegai}{O{i}}{
    \mathbf{\Omega}_{#1}
}
\NewDocumentCommand{\Zi}{O{i}}{
    \mathbf{Z}_{#1}
}
\NewDocumentCommand{\gi}{O{i}}{%
    \mathbf{g}_{#1}
}
\NewDocumentCommand{\Qi}{O{i}}{%
    \mathbf{Q}_{#1}
}
\NewDocumentCommand{\Phii}{O{i}}{%
    \mathbf{\Phi}_{#1}
}
\NewDocumentCommand{\PhiLi}{O{i}}{%
    \mathbf{\Phi}_{L,#1}
}
\NewDocumentCommand{\psii}{O{i}}{%
    \mathbf{\psi}_{#1}
}
\NewDocumentCommand{\deltai}{O{i}}{%
    \mathbf{\delta}_{#1}
}
\NewDocumentCommand{\nSi}{O{i}}{%
    \mathbf{S'}_{#1}
}
\NewDocumentCommand{\conditionijt}{O{i}O{m}O{t}}{%
    S^{(#1,#2)}_{#3}, A^{(#1,#2)}_{#3}
}

\NewDocumentCommand{\tildePhiLijt}{O{i}O{m}O{t}}{%
    \widetilde{\Phi}_{L,#1,#2,#3}
}

\NewDocumentCommand{\tildegijt}{O{i}O{m}O{t}}{%
    \widetilde{g}^{(#1,#2)}_{#3}
}
\NewDocumentCommand{\tildeRijt}{O{i}O{m}O{t}}{%
    \widetilde{R}^{(#1,#2)}_{#3}
}

\NewDocumentCommand{\tildedeltaijt}{O{i}O{m}O{t}}{%
    \widetilde{\delta}^{(#1,#2)}_{#3}
}
\newcommand{\Cov}{\text{Cov}}
\newcommand{\Corr}{\text{Corr}}

\newcommand{\Var}{\text{Var}}

\newcommand{\E}{\mathbb{E}}

\newcommand{\bftheta}{\beta}

\usepackage{xcite}

\makeatletter
\newcommand*{\addFileDependency}[1]{
  \typeout{(#1)}
  \@addtofilelist{#1}
  \IfFileExists{#1}{}{\typeout{No file #1.}}
}
\makeatother

\usepackage{styles}

\begin{document}

\if1\blind
{
\title{\Large{\textbf{Generalized Fitted Q-Iteration with Clustered Data}}}
  \author[1]{Liyuan Hu}
  \author[2]{Jitao Wang}
  \author[2,3]{Zhenke Wu}
  \author[1]{Chengchun Shi\thanks{Corresponding author.}}
  \affil[1]{Department of Statistics, London School of Economics and Political Science, London, UK}
  \affil[2]{Department of Biostatistics, University of Michigan, Ann Arbor, Michigan, USA}
  \affil[3]{Michigan Institute for Data and AI in Society (MIDAS), University of Michigan, Ann Arbor, Michigan, USA}
  \date{\empty}
\maketitle
} \fi

\if0\blind
{
\title{\Large{\textbf{Testing Stationarity and Change Point Detection in Reinforcement Learning}}}
\author{
\bigskip
\vspace{0.5in}
}
\date{}
\maketitle
} \fi

\baselineskip=20pt
\begin{abstract}
This paper focuses on reinforcement learning (RL) with clustered data, which is commonly encountered in healthcare applications. We propose a generalized fitted Q-iteration (FQI) algorithm that incorporates generalized estimating equations into policy learning to handle the intra-cluster correlations. Theoretically, we demonstrate (i) the optimalities of our Q-function and policy estimators when the correlation structure is correctly specified, and (ii) their consistencies when the structure is mis-specified. Empirically, through simulations and analyses of a mobile health dataset, we find the proposed generalized FQI achieves, on average, a half 
reduction in regret compared to the standard FQI.
\end{abstract}

\noindent{\bf Key Words:} 
Reinforcement learning, generalized estimating equations, clustered data.

\baselineskip=22pt

\section{Introduction}
Reinforcement learning (RL) has emerged as a powerful machine learning tool for sequential decision making across various fields. This paper studies RL with cluster-structured data, where subjects are grouped into  clusters, with possible correlations among their observations. 
Our primary motivating application is the Intern Health Study \citep[IHS,][]{necamp2020assessing}, which studied the time-varying causal effect of health-related mobile app prompts on the well-being of first-year medical interns across various institutions in the United States, paving the way for developing more effective and personalized messages tailored for each intern at every decision point. 
These interns naturally form clusters by their memberships to training institutions. Even after controlling for individual-level observed information such as baseline surveys, subjects in the same institution may share unobserved or imperfectly measured factors, e.g., curricula and training environment, and social and peer interactions, resulting in observed intra-institution correlations. 

\begin{figure}[t]
    \centering
    \includegraphics[width=0.4\linewidth]{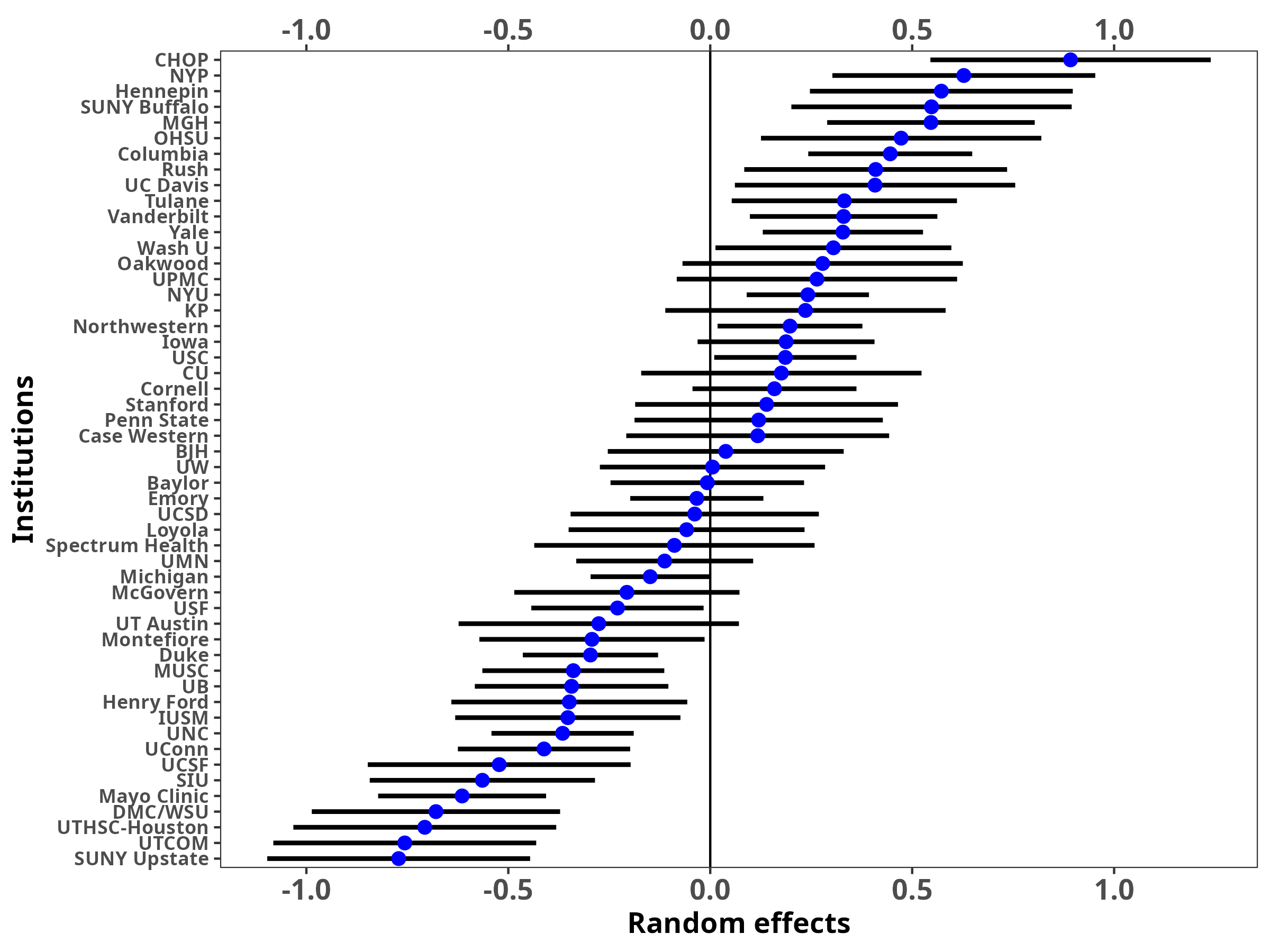}
    \includegraphics[width=0.35\linewidth]{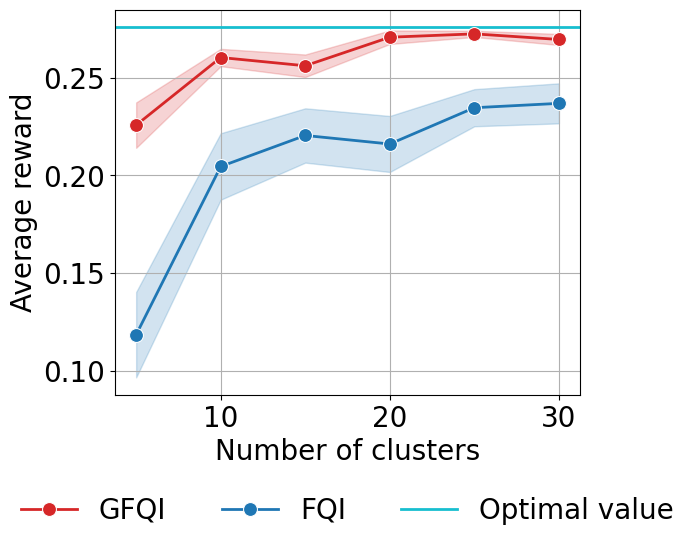}
    \caption{Left panel: Caterpillar plot of random effects for institutions with at least $5$ interns. The error bars indicate 95\% confidence intervals. Right panel: Average reward of policies computed by standard FQI (colored in blue) and the proposed generalized FQI (colored in red) with increasing number of clusters. The horizontal line (colored in cyan) depicts the optimal value, computed by an online deep Q-network (DQN) agent with sufficiently many data. Both the number of subjects per cluster and the time horizon are fixed to $5$.}\label{fig1}
\end{figure}

To illustrate these correlations, we regress the cubic root of interns' daily step counts on the message interventions they received, along with their time-varying observations. We then calculate the resulting residuals and visualize the random intercepts fitted from linear mixed effect models for institutions with at least five individuals in the left panel of Figure~\ref{fig1}, where we observe significantly different random effects across a large number of clusters (institutions), indicating the presence of correlations within these clusters.

The primary challenge of applying RL to clustered data lies in the presence of intra-cluster 
correlations. Existing RL algorithms typically assume that the observations are independent and identically distributed \citep[i.i.d., see e.g.,][]{pmlr-v97-chen19e,pmlr-v120-yang20a} and ignore these dependencies, 
resulting in sub-optimal policy learning. To elaborate the challenges posed by the intra-cluster correlations in policy learning, we provide a numerical example using highly correlated data and visualize the cumulative reward of the estimated optimal policies computed by the traditional fitted Q-iteration \citep[FQI, see e.g.,][]{JMLR:v6:ernst05a,riedmiller2005neural} algorithm and our proposed algorithm (denoted by GFQI) in the right panel of Figure \ref{fig1}. In this example, the ``oracle'' optimal policy achieves a cumulative reward of $0.275$. It can be seen that there is a constant gap between this optimal value and the cumulative reward achieved by the standard FQI. In contrast, the cumulative reward obtained by GFQI, which effectively handles intra-cluster correlations, closely approximates the optimal value.

This work makes several contributions:
\begin{enumerate}[leftmargin=*]
    \item \textbf{Methodologically}, we propose a novel RL algorithm that incorporates generalized estimating equations (GEE) -- a classical statistical tool commonly used in analyzing longitudinal and clustered data \citep{10.1093/biomet/73.1.13, ca1aab99-c8bf-3fed-8f2a-b3d28642c3ed} -- into the FQI for more effective policy learning in the presence of intra-cluster correlations.    
    To our knowledge, this is the first study to explore RL with clustered data that frequently arises in fields such as healthcare, education, and social sciences where subjects are commonly organized in clusters.
    \item \textbf{Theoretically}, we establish the robustness and efficiency of the proposed generalized FQI (referred to as GFQI) algorithm. In particular, we demonstrate that: (i) when the modeled correlation structure is correctly specified, the proposed estimator achieves certain optimality properties; (ii) even when this correlation structure is mis-specified, the proposed estimator maintains consistency, ensuring its robustness \citep{2004Robustness}. These properties apply to both the Q-function estimator and the estimated optimal policy. 
    \item \textbf{Empirically}, we demonstrate the usefulness of our proposal via simulations and real-world data-based semi-synthetic analyses. Our results show that the generalized FQI significantly outperforms the standard correlation-unaware FQI, achieving an approximately $10\%$ reduction in regret under weak intra-cluster correlations and a considerable $80\%$ reduction under strong correlations.
\end{enumerate}

\section{Related works}
We review GEE and RL in this section, as they are closely related to our proposal. 

\subsection{Generalized Estimating Equations}
GEE is a classical statistical tool used for analyzing longitudinal and clustered data. It estimates the parameters of interest by solving a set of estimating equations which specify the conditional mean and variance of a response, while accounting for intra-cluster
correlations via a ``working'' correlation structure 
that is allowed to be mis-specified.

More specifically, suppose we have $n$ clusters, each containing multiple subjects. Let $(X_i^{(j)},Y_i^{(j)})$ denote the predictor-response pair from the $j$th subject within the $i$th cluster. We employ a generalized linear model \citep[GLM,][]{mccullagh2019generalized} to model the relationship between $Y_i^{(j)}$ and $X_i^{(j)}$. Specifically, the conditional mean of $Y_i^{(j)}$, denoted by  $\mu_i^{(j)}=\mathbb{E} (Y_i^{(j)}|X_i^{(j)})$, is modeled  through a link function $g$, such that $g(\mu_i^{(j)})=\beta^\top X_i^{(j)}$, where $\beta$ is a vector of regression coefficients. 

Different from standard GLMs, the responses $Y_i^{(j)}$'s within each cluster are allowed to be correlated, with their correlation structure modelled by a working correlation matrix $\bm{C}_i$. Let $\bm{B}_i$ denote a diagonal matrix whose $j$th diagonal element is given by the conditional variance $\Var(Y_i^{(j)}|X_i^{(j)})$, 
GEE solves the following estimating equations to estimate $\beta$
\begin{equation}\label{eqn:someee}
    \sum_{i=1}^n \frac{\partial \bm{\mu}_i}{\partial \beta}\bm{B}_i^{-1/2} \bm{C}_i^{-1} \bm{B}_i^{-1/2}\left(\bm{Y}_i-\bm{\mu}_i\right)=0,
\end{equation}
where we use boldface letters $\bm{\mu}_i$ and $\bm{Y}_i$ to denote vectors $\{\mu_i^{(j)}\}_j$ and $\{Y_i^{(j)}\}_j$ within the $i$th cluster, respectively. When setting $\bm{C}_i$ to the identity matrix, \eqref{eqn:someee} is reduced to the estimating equations under the i.i.d. assumption. 

GEE has several defining features \citep{10.1093/biomet/73.1.13}:
\begin{enumerate}[leftmargin=*]
    \item Unlike maximum likelihood estimators (MLEs) which require to specify the joint data distribution, GEE only specifies the first two moments. 
    \item When the working correlation structure is correctly specified, the resulting estimator achieves a smaller asymptotic variance compared to MLEs obtained under the i.i.d. assumption.
    \item Even if the correlation structure is mis-specified, GEE still produces consistent and asymptotically normal estimators.
\end{enumerate}
These features make GEE particularly appealing for handling correlated data that frequently arises in biostatistics \citep{10.1093/oso/9780198524847.001.0001,Garrett2008longitudinal,Hedeker2006Longitudinal}. 
Over time, GEE has been extended to accommodate missing data \citep{ Roderick2019Statistical,10.1111/j.1467-9876.2007.00590.x,doi:10.1080/02664760902939604, 10.1111/j.1541-0420.2012.01758.x, Shen2013ModelSO,10.1093/biomet/63.3.581}, high-dimensional data \citep{10.1111/j.1541-0420.2011.01678.x}, and other complex data structures  \citep{https://doi.org/10.1002/sim.4780131106,PARADIS2002175,Liu2009BivariateAA,Friedel2019AMC,Wang2024GeneralizedEE}.

\subsection{Reinforcement Learning} RL has its roots in experimental psychology but has evolved into a powerful AI tool for optimal policy learning through interactions with a given environment \citep{sutton2018reinforcement}. Over the past decade, it has become one of the most popular frontiers in machine learning, with a number of successful applications across diverse fields including game playing \citep{MnihEtAl2015,silver2016mastering}, ride-sharing \citep{xu2018large,tang2019deep,qin2022reinforcement,shi2023dynamic}, real-time bidding \citep{cai2017real,Jin2018RealTimeBW}, finance \citep{Liu2020FinRLAD,Yang2020DeepRL,Wu2020AdaptiveST,carta2021multi} and large language modeling \citep{ouyang2022training,shao2024deepseekmath}. Numerous algorithms have been proposed in the machine learning literature, ranging from the classical Q-learning \citep{watkins1992q} and actor-critic \citep{konda1999actor}, to more advanced trust region policy optimization \citep{pmlr-v37-schulman15}, deep Q-network \citep[DQN,][]{MnihEtAl2015}, quantile DQN \citep{dabney2018distributional}, and more recent offline RL algorithms \citep[see e.g.,][for reviews]{levine2020offline,uehara2022review}. 

In the statistics literature, RL is closely related to a large body of works on learning dynamic treatment regimes  \citep[DTRs, see][for reviews]{chakraborty2013statistical,kosorok2015adaptive,kosorok2019precision,tsiatis2019dynamic,li2023optimal}. More recent works have studied Markov decision processes \citep[MDPs,][]{puterman2014markov} and their variations \citep[e.g., partially observable MPDs][]{krishnamurthy2016partially}, which can broadly be categorized into three types: (i) The first type of methods focuses on the estimation of the optimal policy \citep{ertefaie2018constructing,luckett2020estimating,liao2022batch,li2024settling,zhou2024estimating,yang2022toward,miao2025reinforcement,shi2024statistically,shi2024value,zhong2020risk,zhu2025semi,chen2023steel,shen2025deep,wang2025counterfactually}; (ii) The second type of methods aims to evaluate the return of a target policy \citep{zhou2023distributional,ramprasad2023online,wang2023projected,bian2023off,shi2022statistical,qi2025distributional,shi2024off,zhang2023estimation,kallus2024efficient,liao2021off,luo2024policy,shi2023multiagent,hu2023off,liang2025randomization,sun2024optimal,duan2024optimal,liu2023online}; (iii) The last type of methods considers other related tasks such as variable selection, clustering, order detection or stationarity testing to enable more sample efficient policy learning and/or evaluation   \citet{chen2024reinforcement,li2025testing,ma2023sequential,hu2022doubly,andersen2025graphical}. 

Among all the aforementioned works, our proposal is particularly related to the following two branches of RL algorithms: 
\begin{enumerate}[leftmargin=*]
    \item \textit{\textbf{Fitted Q-iteration}}. FQI is perhaps one of the most popular Q-learning type algorithms. It casts the estimation of the optimal Q-function into a sequential regression problem, enabling the use of existing supervised learning algorithms to estimate the optimal Q-function \citep{JMLR:v6:ernst05a,riedmiller2005neural}. Theoretically, the regret of the resulting estimated optimal policy has been widely studied in the literature  \citep[see e.g.,][]{JMLR:v9:munos08a,pmlr-v97-chen19e,pmlr-v120-yang20a,jin2021pessimism}.
    Our proposal differs from the aforementioned studies both methodologically and theoretically. Methodologically, the proposed GFQI accounts for intra-cluster correlations, unlike standard FQI which treats observations as i.i.d. Theoretically, we conduct a refined analysis to show that our estimated optimal policy incurs a smaller regret than the standard FQI. Such a difference cannot be captured in traditional regret analyses, which typically focus on the order of magnitude of the regret bound, as the two algorithms' regret bounds are of the same order. 
    \item \textbf{\textit{Generalized temporal difference learning}}. 
    Our proposal is also closely related to the generalized temporal difference (GTD) learning algorithm developed by \citet{Ueno2011GeneralizedTL}, which proposes to utilize a set of estimating equations to solve temporal difference learning for policy evaluation and derives an ``optimal'' estimating equation that minimizes the variance of the estimated parameters within that set. 
    A similar work by \citet{NEURIPS2021_3e6260b8} develops a variance-aware weighted regression method that assigns weights inversely proportional to the variance of the value function for parameter estimation and the subsequent policy evaluation. However,  utilizing the variance alone without modifying the estimating equation is insufficient to achieve semi-parametric efficiency \citep{Ueno2011GeneralizedTL}. 
    Our proposal differs from GTD in that: (i) While both our proposal and GTD identify optimal estimating equations to minimize the variance of the parameter estimates, we employ GEE to account for the intra-cluster correlations, whereas GTD still treats observations as i.i.d. (ii) We consider policy learning, a more complicated problem than policy evaluation which often serves as an intermediate step in policy learning \citep{sutton2018reinforcement}. 
\end{enumerate}

\section{Clustered Markov Decision Processes}\label{sec:statsframework}
We propose a clustered MDP to model the intra-cluster correlations within the clustered data. To elaborate this DGP, we discuss its connections to and differences from the standard MDP model in this section. Notably, the standard MDP requires the error residuals to be independent. In contrast, the proposed clustered MDP allows residuals to be dependent over  
population.

We first introduce the standard MDP commonly employed in RL to formulate numerous sequential decision making problems. 
An MDP is typically denoted by $\langle\mathcal{S}, \mathcal{A}, \mathcal{T}, \mathcal{R}, \rho, \gamma\rangle$ where $\mathcal{S}$ and $\mathcal{A}$ denote the state and action spaces, respectively; $\mathcal{T}$ and $\mathcal{R}$ are the state transition and reward functions;  $\rho$ denotes the initial state distribution and  $\gamma$ denotes the discount factor, bounded between $0$ and $1$. 

\begin{figure}[t]
    \centering
    \includegraphics[width=0.35\linewidth]{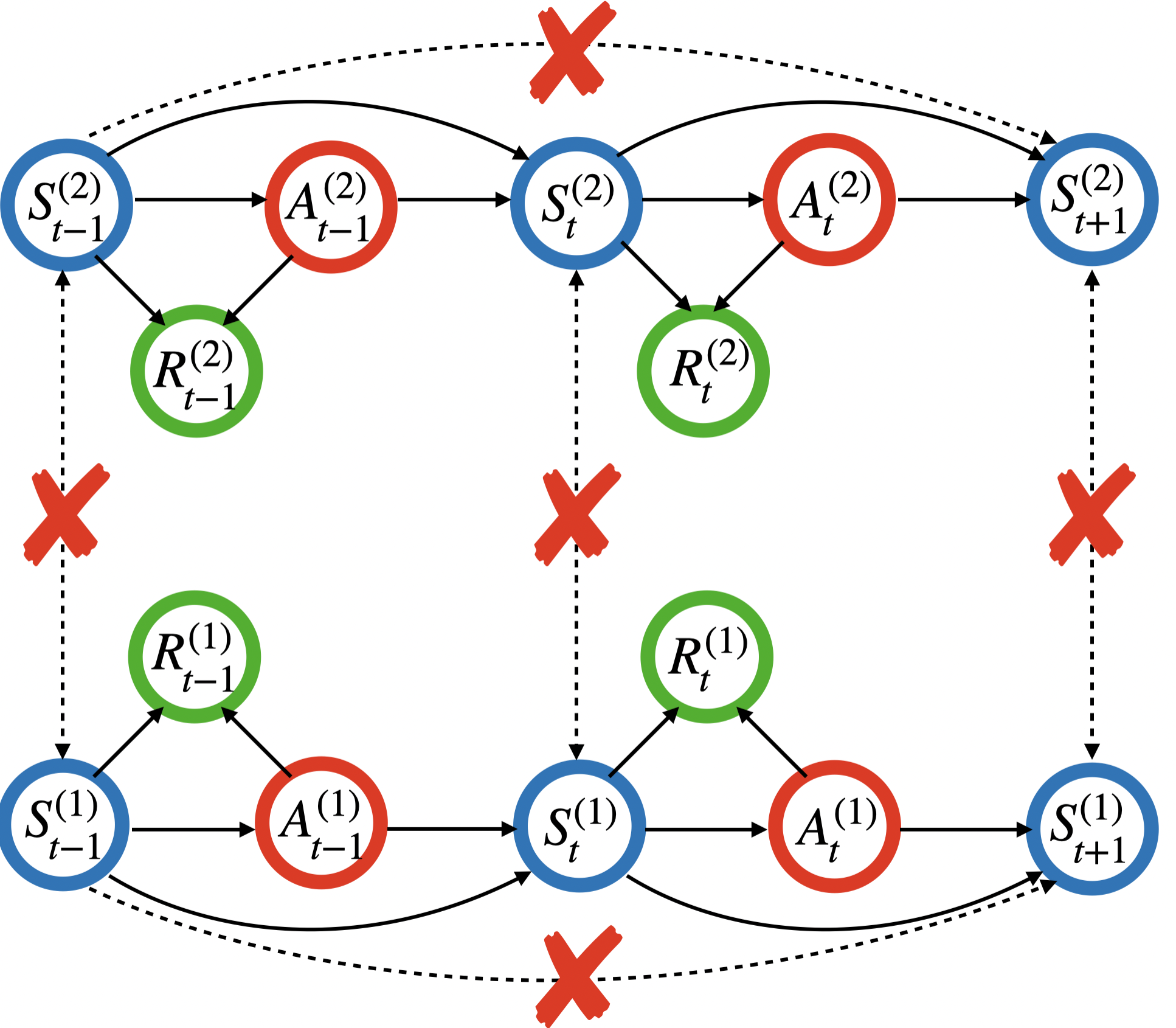}
    \includegraphics[width=0.4\linewidth]{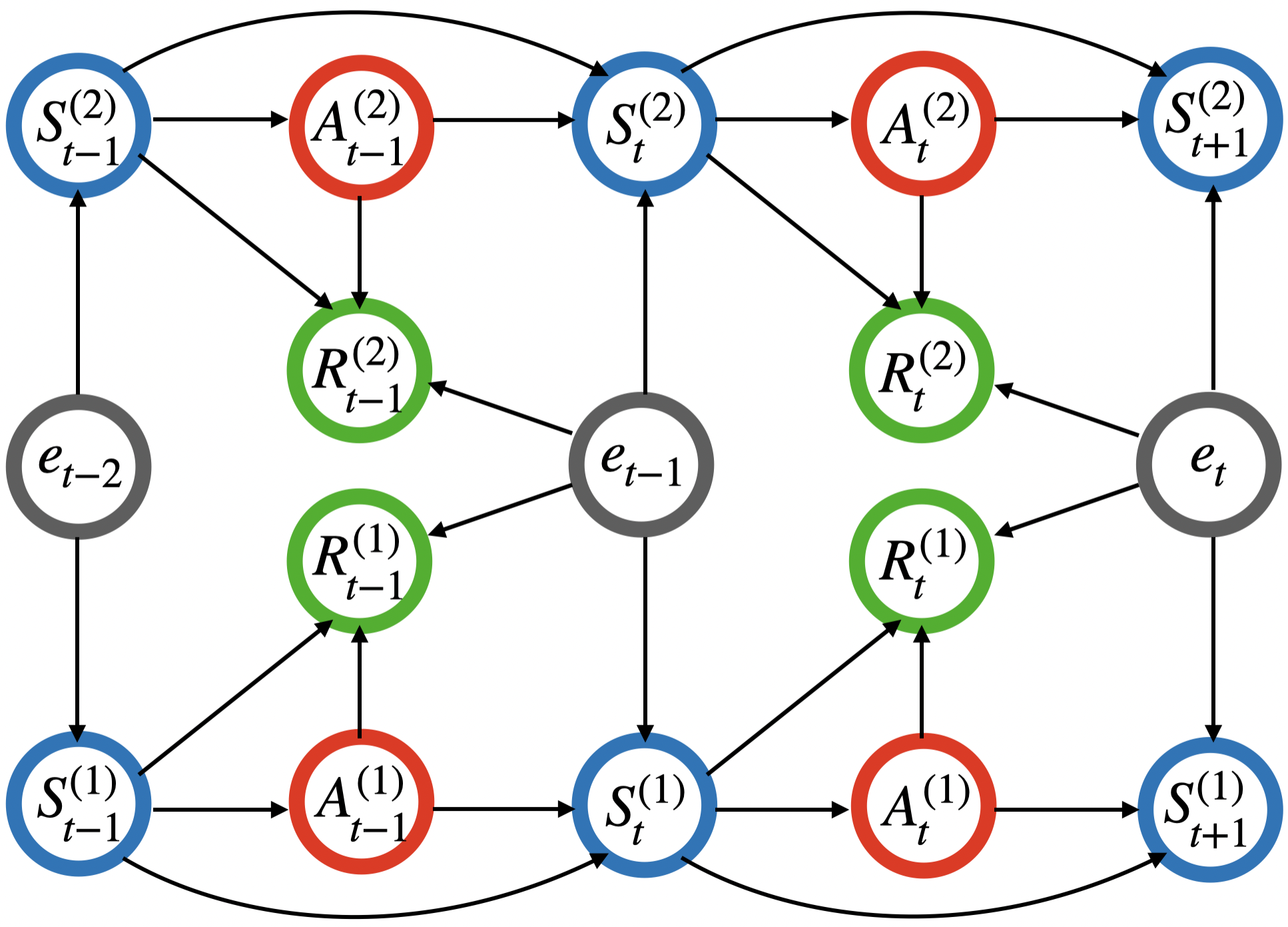}
    \caption{Left panel: Visualization of MDPs. $(S_t^{(1)}, A_t^{(1)},R_t^{(1)})_{t\ge 1}$ and $(S_t^{(2)}, A_t^{(2)},R_t^{(2)})_{t\ge 1}$ denote two data trajectories. The Markov assumption precludes paths $S_{t-1}^{(1)}\to S_{t+1}^{(1)}$ and $S_{t-1}^{(2)}\to S_{t+1}^{(2)}$. The independence assumption precludes paths across the two trajectories. Right panel: A clustered MDP example.}
    \label{fig2model}
\end{figure}

Let $(S_t,A_t,R_t)_{t\ge 0}$ denote the sequence of state-action-reward triplets generated under this MDP. Its data generating process can be described as follows:
\begin{enumerate}[leftmargin=*]
    \item At the initial time $t=0$, the environment's state $S_t\in \mathcal{S}$ is generated according to $\rho$, i.e., $S_0\sim \rho(\bullet)$;
    \item Subsequently, at each time $t$, the agent selects an action $A_t\in \mathcal{A}$ after observing $S_t$; 
    \item Next, the environment provides an immediate reward $R_t$ with an expected value of $\mathcal{R}(A_t,S_t)$, and transits into a new state $S_{t+1}\sim \mathcal{T}(\bullet|A_t,S_t)$ at time $t+1$;
    \item Steps 2 and 3 repeat until the environment reaches a terminal state. 
\end{enumerate}
An MDP contains two basic assumptions (see the left panel of Figure \ref{fig2model} for illustrations): 
\begin{enumerate}[leftmargin=*]
    \item[(i)] The first one is a Markov assumption, which requires both immediate reward and future state to depend on the past data history only through the current state-action pair. This assumption precludes any directed path from the past state to future state that does not traverse current state-action pair.
    \item[(ii)] The second one is an independence assumption, which requires the state-action-reward triplets to be independent across different trajectories. This assumption precludes any directed path among trajectories.
\end{enumerate}
The goal is to learn an optimal policy $\pi^*$ that maximizes the expected cumulative reward $\sum_{t\ge 0} \gamma^t \mathbb{E}^{\pi}(R_t)$ where $\mathbb{E}^{\pi}(R_t)$ denotes the expected reward the agent receives at time $t$ following a given policy $\pi$.

This paper studies MDPs under the clustered data setting. We use $\langle\mathcal{S},\mathcal{A},\mathcal{M}, \{\mathcal{T}_m\}_{m},\mathcal{R},\{\rho_m\}_m,\gamma\rangle$ to denote such a clustered MDP. In this notation:
\begin{itemize}[leftmargin=*]
    \item $\mathcal{S}$, $\mathcal{A}$, $\mathcal{R}$ and $\gamma$ denote the same state space, action space, reward function and discount factor as those in a standard MDP;
    \item $\mathcal{M}$ denotes the probability mass function of $M$ -- the number of subjects within each cluster; 
    \item $\rho_M$ and $\mathcal{T}_M$ denote the joint distribution function of all $M$ initial states, and the transition function of all the $M$ states, assuming there are $M$ subjects in the cluster. When restricting to a single state, its marginal initial state distribution and marginal transition function are reduced to $\rho$ and $\mathcal{T}$, respectively.
\end{itemize}
Such a clustered MDP generates $M$ trajectories, each containing a sequence of state-action-reward triplets $(S_t^{(m)}, A_t^{(m)}, R_t^{(m)})_{t\ge 0}$ over time, for $1\le m\le M$. 
Similar to the MDP model, each trajectory marginally follows an MDP  $\langle\mathcal{S},\mathcal{A},\mathcal{T},\mathcal{R},\rho,\gamma\rangle$.
However, different from the MDP, these  trajectories can be dependent.

To elaborate, we visualize a clustered MDP in the right panel of Figure \ref{fig2model}. In this example, a common residual process $\{e_t\}_{t\ge 1}$ affects the rewards and/or states of two subjects  simultaneously. For instance, the reward residuals of two subjects could be identical, 
$$R_t^{(1)}-\mathcal{R}(A_t^{(1)},S_t^{(1)})=R_t^{(2)}-\mathcal{R}(A_t^{(2)},S_t^{(2)}),\,\,\forall t.$$
This invalidates the independence assumption. However, we do require $\{e_t\}_{t\ge 1}$ to be independent over time to meet the Markov assumption. Otherwise, each trajectory follows a partially observable MDP, invalidating most RL algorithms that rely on the MDP assumption. 
As such, each trajectory in a clustered MDP follows an MDP marginally, and its expected cumulative reward $\sum_{t\ge 0} \gamma^t \mathbb{E}^{\pi}(R_t)$ under any policy $\pi$\footnote{In this paper, we focus on policies whose action assignment mechanisms depend only on their own trajectory's data history and not on other trajectories.} remains the same as if it were generated independently from an MDP. Consequently, $\pi^*$ continues to maximize the expected cumulative reward. Moreover, most existing RL algorithms remain consistent to identify $\pi^*$. 
However, they are no longer as sample efficient as under Model I, as demonstrated in the right panel of Figure \ref{fig1}.

\section{Generalized Fitted Q-iteration}\label{sec:method}
In this section, we proposed a generalized FQI algorithm for more effective policy learning in the presence of intra-cluster correlations within a clustered MDP. For illustration purposes, we begin by introducing the standard FQI algorithm and adapting GTD for policy learning (denoted as AGTD). However, neither FQI nor AGTD addresses data correlations. Our proposal, presented later, is built upon these algorithms and specifically designed to manage such correlations. All the three algorithms are Q-learning type methods, which focus on estimating the following optimal Q-function, defined by
\begin{equation*}
    Q^*(a,s)=\sum_{t\ge 0} \gamma^t \mathbb{E}^{\pi^*}(R_t|A_0=a,S_0=s),
\end{equation*}
and derive the estimated optimal policy as the greedy policy with respect to the estimated Q-function. See Algorithms \ref{alg:fqi}--\ref{alg:optimal_fqi} for their pseudocodes. 

\smallskip

\noindent \textbf{Algorithm I (FQI)}. As commented in Section \ref{sec:statsframework}, standard FQI remains consistent in identifying the optimal policy at a population level when applied to clustered MDPs, although it lacks sample efficiency in finite samples. 
It organizes all data into a collection of state-action-reward-next-state $(S,A,R,S')$ tuples $\mathcal{D}$ and iteratively estimates $Q^*$ via supervised learning. More specifically, the initial Q-function estimator $Q^{(0)}$ is set to a zero function. During the $k$th iteration, FQI solves the following optimization,
\begin{equation}\label{eqn:FQI}
    \argmin_{Q\in \mathcal{Q}} \sum_{\substack{(S,A,R,S')\in \mathcal{D}}} \Big[R+\max_a\gamma Q^{(k-1)}(a,S') - Q(A,S)\Big]^2,
\end{equation}
to compute $Q^{(k)}$. 

Here, $\mathcal{Q}$ denotes the function class used to model $Q^*$. It can be any parametric class, such as the class of linear models where $\mathcal{Q}=\{Q: Q(a,s)=\beta^\top \phi(a,s)\}$ for a set of state-action features $\phi$ with dimension $d$, or other more complex models such as neural networks.

\begin{algorithm}[t]
\caption{Fitted Q-iteration (FQI)}\label{alg:fqi}
\begin{algorithmic}[1]
\State Initialize $Q^{(0)}\gets 0$ and $k \gets 0$.
\Repeat
    \State $k \gets k + 1$
    \State Update $Q^{(k)}$ by solving 
    \eqref{eqn:FQI}.
\Until{Convergence criteria is met.}
\State \Return $Q^{(k)}$
\end{algorithmic}
\end{algorithm}
\begin{algorithm}[t]
\caption{Adapted Generalized TD (AGTD)}\label{al:agtd}
\begin{algorithmic}[1]
    \State Initialize $\bftheta^{(0)}\gets 0$ and $k \gets 0$.
\Repeat
    \State $k \gets k + 1$
    \State Estimate $\pi^*$, the expectation in \eqref{eqn:Phi*} and $\sigma^2$ by the currently estimated optimal policy and supervised learning to estimate $\phi^*$ and $\sigma^2$. Let $\widehat{\phi}^*$ and $\widehat{\sigma}^2$ denote their estimators. 
    \State Update $\bftheta^{(k+1)}$ by solving 
    $$
    \sum_{(S,A,R,S')\in \mathcal{D}} \frac{\widehat{\phi}^*(A,S)}{\widehat{\sigma}^{2}(A,S)}\Big[R+\gamma \times \max_a\phi^\top(a,S')\beta^{(k)}- \phi^\top(A,S)\beta^{(k+1)}\Big]=0.
    $$
\Until{Convergence criteria is met.}
\State \Return $\bftheta^{(k)}$
\end{algorithmic}
\end{algorithm}

\smallskip

\noindent \textbf{Algorithm II (Adaptation of GTD for policy learning)}. GTD is originally designed for the purpose of Q-function evaluation. Here, we extend it to learn the optimal policy $\pi^*$. Following GTD, we assume the optimal Q-function is linear, i.e., $Q^*(a,s)=\phi^\top(a,s) \beta^*$ for some $\beta^*\in \mathbb{R}^d$. According to the Bellman optimality equation \citep[see e.g.,][]{agarwal2019reinforcement}, for any $d$-dimensional state-action feature $\Phi$, the following estimating equation is unbiased for estimating $\beta^*$, 
\begin{eqnarray}\label{eqn:GTDee}
    \sum_{(S,A,R,S')\in \mathcal{D}} \Phi(A,S)\Big[R+\gamma\max_a\phi^\top(a,S')\widehat{\beta} - \phi^\top(A,S)\widehat{\beta}\Big]=0,
\end{eqnarray}
i.e., the left-hand-side (LHS) of \eqref{eqn:GTDee} equals zero when replacing $\widehat{\beta}$ with $\beta^*$. This motivates us to solve the above equation to compute the estimator $\widehat{\beta}$ for $\beta^*$.
 
The choice of the state-action feature $\Phi$ in \eqref{eqn:GTDee} is crucial. Each $\Phi$ leads to a different estimator $\widehat{\beta}$. Motivated by GTD, we aim to identify the optimal state-action feature that minimizes the mean squared error (MSE) of $\widehat{\beta}$. Assuming the data comes from a standard MDP with independent trajectories, it follows from Theorem \ref{theo:asym properties} (see Section \ref{sec:theo}) that the resulting optimal $\Phi$ is specified by $\phi^*(a,s)\sigma^{-2}(a,s)$ where
\begin{equation}\label{eqn:Phi*}
    \phi^*(A,S)=\phi(A,S)-\gamma\mathbb{E} [\phi(\pi^*(S'),S')|A,S],
\end{equation}
and $\sigma^2(A,S)$ denotes the conditional variance of the temporal difference (TD) error $R+\gamma \max_a Q^*(\pi^*(S'),S')-Q^*(A,S)$ given $A$ and $S$. 

To solve this estimating equation, we estimate $\beta^*$ iteratively in a similar manner to FQI. At each iteration, we first use the currently estimated optimal policy -- greedy with respect to the current Q-function estimator -- and supervised learning to approximate $\pi^*$, the expectation in \eqref{eqn:Phi*} and $\sigma^2$, in order to construct the optimal $\Phi$. We next solve the resulting estimating equation in \eqref{eqn:GTDee}. 
The aforementioned procedure is repeated until convergence. 

\smallskip

\noindent \textbf{Algorithm III (GFQI)}. Algorithm II derives the optimal basis function $\phi^*$ under the independence assumption, which is clearly violated in clustered MDPs. To address this limitation, we incorporate GEE into FQI to effectively accommodate data correlations. 

For a clustered MDP, 
we use boldface letters $\bfS$, $\bfA$, $\bfR$ and $\bfS'$ to denote the collections of state-action-reward-next-state tuples across all $M$ subjects within a cluster at a given time, i.e., $\bfS=(S^{(1)\top}, \cdots, S^{(M)\top})^\top$ and $\bfA$, $\bfR$, $\bfS'$ can be defined similarly. 
Under the given model assumption, the parameter of interest $\bftheta^*$ satisfies the Bellman optimality equation $\mathbb{E} [\bm{\delta}(\bfS,\bfA,\bfR,\bfS';\bftheta^*)]=\bm{0}$ where $\bm{\delta}(\bfS,\bfA,\bfR,\bfS';\bftheta^*)$ denotes an $M$-dimensional cluster-wise TD error whose $j$th entry is given by
\begin{align*}
R^{(j)} + \max_{a} \gamma  \phi(a, S^{(j)'})\bftheta^*-  \phi(A^{(j)}, S^{(j)})\bftheta^*. 
\end{align*}
This motivates us to construct the following estimating equation to estimate $\beta^*$:
\begin{equation}\label{eq:optimal bellman}
    \sum_{(\bfS,\bfA,\bfR,\bfS')} \bm{\Phi}(\bfA,\bfS)\bm{\delta}(\bfS,\bfA,\bfR,\bfS';\widehat{\bftheta})=\bm{0},
\end{equation}
for any $d\times M$ matrix $\bm{\Phi}$ that is a function of $\bfA$ and $\bfS$, where the summation in \eqref{eq:optimal bellman} is taken over all $(\bfS,\bfA,\bfR,\bfS')$ tuples across different times and clusters. 

According to Theorem \ref{theo:asym properties}, the optimal $\bm{\Phi}$ is given by
\begin{equation*}
    \bm{\Phi}^*(\bfA,\bfS)=\Big[\phi^*(A^{(1)},S^{(1)}),\cdots, \phi^*(A^{(M)},S^{(M)})\Big] \bfV^{-1},
\end{equation*}
where $\phi^*$ is defined in \eqref{eqn:Phi*}, and $\bfV$ 
is the covariance matrix of the cluster-wise TD error. 

Similar to GEE, we model $\bfV$ via $\bm{B} \bm{C} \bm{B}$ where $\bm{B}$ is a diagonal matrix with each $j$th diagonal element equal to the standard deviation of the $j$th TD error $\sigma(A^{(j)},S^{(j)})$. The matrix $\bm{C}$ in the middle serves as a working correlation matrix to model the intra-cluster correlations. Typically, $\bm{C}$ is set to an exchangeable correlation matrix, where the diagonal elements are $1$ and the off-diagonal elements equal some correlation coefficient $-1<\rho<1$.

To solve the resulting generalized estimating equation in \eqref{eq:optimal bellman}, we employ the same iterative approach to AGTD. Specifically, at each iteration, we estimate the correlation coefficient $\rho$ and $\phi^*$, and update the parameter $\beta^{(k)}$ by solving
\begin{equation}\label{eq:optimal FQI iteration}
\begin{split}
    \sum_{(\bfS, \bfA,\bfR,\bfS^\prime)}
    \bm{\Phi}^{*}(\bfA,\bfS) 
    \bm{\delta}(\bfS,\bfA,\bfR,\bfS';\bftheta^{(k)}, \bftheta^{(k-1)})=0,
\end{split}
\end{equation}
where $\bm{\delta}(\bfS,\bfA,\bfR,\bfS';\bftheta^{(k)}, \bftheta^{(k-1)})$ denotes the TD error vector with the $j$th entry given by
\begin{equation*}
    R^{(j)} + \max_{a} \gamma  \phi(a, S^{(j)'})\bftheta^{(k-1)}-  \phi(A^{(j)}, S^{(j)})\bftheta^{(k)}.
\end{equation*}
Since we iteratively solve the GEE to estimate the optimal Q-function, we refer to our algorithm as generalized FQI, or GFQI, for short. Notice that GFQI is reduced to AGTD when the correlation matrix $\bm{C}$ is set to the identity matrix. 

\addtocounter{algorithm}{1}
  
\begin{algorithm}[t]
\caption{Generalized Fitted Q-iteration (GFQI)}\label{alg:optimal_fqi}
\begin{algorithmic}[1]
\State Set the iteration counter $k \gets 0$.
\State Initialize $\bftheta^{(0)}\gets 0$ so that the Q-function $Q^{(k)}(a,s)=\phi^\top(a,s)\beta^{(k)}$ is initialized to zero. 
\Repeat
    \State Estimate $\pi^*$ by the greedy policy $\pi^{(k)}\gets \arg\max_a \phi^\top(a,s)\beta^{(k)}$ with respect to $Q^{(k)}$.
    \State Estimate $\phi^*(A,S)=\E[\phi(\pi^*(S^\prime), S^\prime)| A,S]$ using supervised learning, with $\{\phi(\pi^{(k)}(S^\prime), S^\prime)\}$ as the response variables and $\{(A,S)\}$ as the predictors. 
    \State Calculate the TD error $$\delta(A^{(m)}_t,S^{(m)}_t,R^{(m)}_t,S_{t+1}^{(m)})\gets R^{(m)}_t+\gamma \max_a Q^{(k-1)}(a, S_{t+1}^{(m)})-Q^{(k)}(A_t^{(m)},S_t^{(m)})$$ for $m=1,\cdots,M$ and $t=1,\cdots,T$. 
    \State Use these TD errors to estimate $\bfV=\{V_{m_1,m_2}\}_{m_1,m_2}$ where $$V_{m_1,m_2}=\Cov(\delta(A^{(m_1)}_t,S^{(m_1)}_t,R^{(m_1)}_t,S_{t+1}^{(m_1)}),\delta(A^{(m_2)}_t,S^{(m_2)}_t,R^{(m_2)}_t,S_{t+1}^{(m_2)})).$$
    \State Estimate $\bm{\Phi}^*$ by plugging the estimated $\phi^*$ and $\bm{V}$ into
    \begin{equation*}
        \bm{\Phi}^*(\bfA,\bfS)=\Big[\phi^*(A^{(1)},S^{(1)}),\cdots, \phi^*(A^{(M)},S^{(M)})\Big] \bfV^{-1},
    \end{equation*}
    \State Update $\bftheta^{(k+1)}$ by solving 
    \begin{eqnarray*}
        \sum_{(\bfS, \bfA,\bfR,\bfS^\prime)}
    \bm{\Phi}^{*}(\bfA,\bfS) 
    \bm{\delta}(\bfS,\bfA,\bfR,\bfS';\bftheta^{(k)}, \bftheta^{(k-1)})=0,
    \end{eqnarray*}
    with the $j$th element entry of $\bm{\delta}(\bfS,\bfA,\bfR,\bfS';\bftheta^{(k)}, \bftheta^{(k-1)})$ given by $\sigma(A^{(j)},S^{(j)})$.
    \State $k \gets k + 1$
\Until{Convergence criteria is met.}
\State \Return $\bftheta^{(k)}$
\end{algorithmic}
\end{algorithm}

\section{Statistical Properties}\label{sec:theo}
In this section, we study the theoretical properties of the proposed GFQI (Algorithm \ref{alg:optimal_fqi}). We first obtain a bias-variance decomposition of the MSE of the estimated $\beta^*$. We next analyze the regret of the resulting estimated optimal policy. See Table \ref{tab:theory} for a summary of our theoretical findings. Without loss of generality, we assume that the numbers of subjects for all clusters are the same and denote this number as $M$. Let $N$ denotes the number of state-action-reward-next-state tuples $(\bfS,\bfA,\bfR,\bfS')$ in the dataset.

We impose Assumptions \ref{as:bounded reward} -- \ref{as:Q third derivatives} to derive our theories and relegate them to the Appendix to save space. We briefly discuss these assumptions below: 
\begin{itemize}[leftmargin=*]
    \item Assumptions \ref{as:bounded reward}, \ref{as:linear mdp} and \ref{as:completeness} are standard in the RL literature \citep[see e.g.,][]{pmlr-v97-chen19e, pmlr-v120-yang20a, uehara2022finitesampleanalysisminimax}. They require that all immediate rewards be bounded by a constant $R_{\max} < \infty$, the linear function class be rich enough to cover the reward function $\mathcal{R}$ (realizability), and be closed under the Bellman operator (completeness), respectively. 
    \item Assumption \ref{as:ergodic}  requires the underlying clustered MDP to satisfy the geometric ergodicity condition \citep{bradley2005basic}, which relaxes the existing i.i.d. data assumption commonly imposed in RL \citep[see e.g.,][]{pmlr-v97-chen19e,pmlr-v120-yang20a,dai2020coindice}. It is equivalent to the $\beta$-mixing condition \citep{luckett2020estimating} and is strictly weaker than the uniform ergodicity assumption imposed in the literature \citep{zou2019finite}.
    \item {\color{black}Assumption \ref{as:bounded Dphi -gamma Dphi'} is similar to the stability assumption used in the off-policy evaluation literature \citep[see e.g.,][]{perdomo2022complete}, but is specifically tailored for policy learning. This assumption is frequently employed in settings where linear function approximation is utilized for modeling the (optimal) Q-function \citep{ertefaie2018constructing,shi2022statistical,perdomo2023complete,bian2023off,li2025testing}. Following the arguments in \citet{shi2022statistical}, we can show that this assumption is likely to hold when data are collected using the $\epsilon$-greedy algorithm; see Section \ref{sec:supp:ass} of the Supporting Information for details.}
    \item Assumption \ref{as:noofiterations} requires the number of GFQI iterations $K$ to be much larger than $\log(N)/\log(\gamma^{-1})$. This assumption is mild as $K$ is user-specified. 
    \item Finally, Assumption \ref{as:unique optimal pi} requires the uniqueness of the optimal policy whereas Assumption \ref{as:Q third derivatives} requires the value function to be a smooth function of the model parameter $\beta$. These assumptions enable us to derive the leading terms of the MSE and regret, which are essential for comparing the performance of different algorithms; see our discussions below Theorems \ref{theo:asym properties} and \ref{theo:regret}. 
\end{itemize}

\begin{table}[t]
    \centering
    \begin{tabular}{c|c|c}\toprule
        Correlation matrix & Theorem 1 (estimated $\beta^*$)  & Theorem 2 (policy)  \\ \midrule
        Correctly specified & minimal asymptotic variance & minimal asymptotic regret \\ \midrule
        
        Mis-specified &  consistent & consistent\\ \bottomrule
    \end{tabular}
    \caption{A summary of our theoretical findings.}
    \label{tab:theory}
\end{table}

\subsection{Error analysis of the estimated $\beta^*$}\label{eqn:asynormal}
The following theorem establishes the efficiency and robustness of the estimated $\beta^*$ obtained from Algorithm \ref{alg:optimal_fqi}. 

\begin{theorem}[MSE]\label{theo:asym properties}
Suppose Assumptions \ref{as:bounded reward}-\ref{as:unique optimal pi} are satisfied. For a sufficiently large sample size $N$,  $\widehat{\bftheta}$ computed by Algorithm \ref{alg:optimal_fqi} attains the following properties:
\begin{enumerate}[leftmargin=*]
\item \textbf{Robustness}: The MSE of $\widehat{\bftheta}$ is given by
\begin{eqnarray}\label{eqn:errorbound}
    \frac{\textrm{tr}[\bm{W}^{-1}(\bm{\Phi})\bm{\Sigma}(\bm{\Phi}) (\bm{W}^{-1}(\bm{\Phi}))^\top]}{N}
    +O\Big(\frac{R_{\max}\log(N)}{N^{\frac{3}{2}}(1-\gamma)^{5}}\Big), 
\end{eqnarray}
where tr$(\bullet)$ denotes the trace of a matrix, $\bm{W}(\bm{\Phi})=(1-\gamma)^{-1}M^{-1}\E\left[ \bm{\Phi}(\bfA,\bfS)\left\{\bm{\phi}(\bfA, \bfS)- \gamma\bm{\phi}(\pi^{*}(\bfS^\prime), \bfS^\prime)\right\}\right]$, $\bm{\phi}(\bm{A},\bm{S})=[\phi(A^{(1)},S^{(1)}),\cdots, \phi(A^{(M)},S^{(M)})]$ and $\bm{\Sigma}(\bm{\Phi}) =M^{-1}\E[\bm{\Phi}(\bfA,\bfS)\bfV\bm{\Phi}^\top(\bfA,\bfS)]$.

\item \textbf{Efficiency}:  
$\widehat{\bftheta}$ reaches the minimal asymptotic MSE -- obtained by removing errors that are high-order in the sample size $N$ -- among the class of solutions to \eqref{eq:optimal bellman} if $\bm{\Phi} = \bm{\Phi}^*$ with a correctly specified correlation matrix $\bm{C}$.
    \end{enumerate}
\end{theorem}
The first part of Theorem \ref{theo:asym properties} derives a rigorous bias-variance decomposition for the MSE of $\widehat{\beta}$. Specifically, the first term in the error bound (see \eqref{eqn:errorbound}) represents the leading asymptotic variance, which scales as $N^{-1}$, whereas the second term arises from its finite-sample bias, which decays at a faster rate of $N^{-3/2}$. As a result, the bias term diminishes more rapidly than the asymptotic variance as the sample size $N$ increases. Meanwhile, the bias term is proportional to the reward upper bound $R_{\max}$ and increases with the $(1-\gamma)^{-1}$ term, which can be interpreted as the ``horizon'' in episodic tasks. Finally, regardless of the working correlation structure employed, both the bias term and asymptotic variance decay to zero as $N$ approaches infinity, demonstrating the robustness  of our proposed GFQI.

The second part of Theorem \ref{theo:asym properties} demonstrates that, when setting $\bm{\Phi}$ to $\bm{\Phi}^*$ with a correctly specified correlation matrix, the proposed GFQI minimizes the leading asymptotic variance of the resulting estimator. This efficiency is achieved through the following three key steps:  (i) using $\phi^*$ instead of $\phi$ itself to construct the state-action feature; (ii) inversely weighting each state-action feature by the conditional variance of the TD error $\sigma^2$; (iii) incorporating $\bm{C}$ to account for the within-cluster correlation. In comparison:
\begin{itemize}[leftmargin=*]
    \item \textit{\textbf{Standard FQI}} adopts none of these steps, which often incurs a large MSE;
    \item \textit{\textbf{Variance-aware regression}} \citep{NEURIPS2021_3e6260b8} functions similarly to the second step but remains inefficient as it omits the first and third steps;
    \item \textit{\textbf{AGTD}} improves upon variance-aware regression by incorporating the first two steps, but still falls short of efficiency because it does not exploit the correlation structure. 
\end{itemize}
Our work builds on AGTD by accounting for this correlation, resulting in a more efficient estimator.

Finally, we emphasize that a major difference of our proposal from the existing literature lies in the refined theoretical analysis of the error terms. While existing results in the machine learning literature often focus on establishing the order of the MSE, our analysis goes further by explicitly characterizing the leading term and developing a method to optimize it. Without such an analysis, all the aforementioned methods would achieve the same order of magnitude, making it impossible to distinguish their performance.

\subsection{Regret analysis}
The following theorem demonstrates that the estimation error of $\widehat{\beta}$ can be translated into the regret of the resulting estimated optimal policy. 

\begin{theorem}\label{theo:regret}
    Suppose Assumptions \ref{as:bounded reward}-\ref{as:Q third derivatives} are satisfied. For a sufficiently large $N$, we have the following results:
    \begin{enumerate}[leftmargin=*]
        \item \textbf{Robustness}: 
    The regret of the estimated optimal
policy is given by
\begin{equation}
\begin{split}
    -\frac{1}{2}\mathrm{tr}(\Var_A(\widehat{\bftheta})H)+ 
    O\Big(\frac{R_{\max}^{3/2}\log(N)}{(1-\gamma)^6 N^{\frac{3}{2}}}\Big),
\end{split}
    \label{eqn:regretbound}
\end{equation}
where $\Var_{A}(\widehat{\bftheta})$ denotes the asymptotic covariance matrix of $\widehat{\bftheta}$, given by $N^{-1}\bm{W}^{-1}(\bm{\Phi}) \bm{\Sigma}(\bm{\Phi})(\bm{W}^{-1}(\bm{\Phi}))^\top$, and
{\color{black}$H$ denotes the Hessian matrix 
\begin{eqnarray*}
    H=\frac{\partial \mathcal{V}(\beta)}{\partial \beta \partial \beta^\top}\Big|_{\beta=\beta^*},
\end{eqnarray*}
where $\mathcal{V}(\beta)$ denotes the expected cumulative reward $\sum_{t\ge 0} \gamma^t\E^{\pi(\beta)}(R_t)$ under the greedy policy $\pi_\beta(s)=\arg\max_a \phi^\top(a,s)\beta$ parameterized by $\beta$.}

        \item \textbf{Efficiency}: The regret bound in \eqref{eqn:regretbound} is asymptotically minimized {\color{black}(i.e., its leading first term is minimized)} when the correlation matrix is correctly specified. 
    \end{enumerate}

\end{theorem}

As commented earlier, existing theoretical analyses provide only the order of magnitude of the policy regret, which is insufficient for comparing different algorithms when their regrets share the same order. To address this limitation, similar to Theorem \ref{theo:asym properties}, we refine the existing analysis by decomposing the regret into a leading term, which is related to the asymptotic variance of $\widehat{\bftheta}$, and a higher-order residual term in \eqref{eqn:regretbound}. {\color{black}By definition, $\mathcal{V}(\beta)$ is maximized at $\beta^*$. As such, $H$ is negative semidefinite, making the leading term  nonnegative.} Given that the leading term is of the order $O(N^{-1})$, it is indeed the dominating factor. 
Meanwhile, this dominating factor is minimized when the correlation matrix is correctly specified, which demonstrates the efficiency of the proposed GFQI. When misspecified, both the leading term and the higher-order term converge to zero as the sample size $N$ approaches infinity, verifying the robustness property.

\section{Numerical Study}
\label{sec:numirical}

In this section, we 
employ the IHS dataset to create a simulation environment to investigate the empirical performance of the proposed GFQI in real-world applications.
To save space, 
additional synthetic data analyses results
are relegated to Section \ref{appendix:sec:simu} of the Supplementary Materials. 
The code for reproducing these experiments is available at \url{https://github.com/zaza0209/GEERL}.

To mimic the data generation process in the IHS study, we first train a state transition model and a reward model. Here, the state variable $S_{t}$ is set to be the cubic-root of daily step count of an intern at day $t-1$ and the reward variable $R_{t}$ is set to be their daily mood score at day $t$. {\color{black}We remark that the original daily step count is heavy-tailed, and the cubic-root transformation brings its distribution closer to normality. See Figure C.5 of \citet{Li2025supplement} for a visualization of the resulting transformed variable. }The action $A_{t}$ is binary, where $A_{t} = 1$ indicates that a message was received on day $t$, and $A_t = 0$ indicates no message was received. Further details regarding the generative model and correlation structure are provided in Section~\ref{appendix:sec:simu} of the Supporting Information.

The sample size is determined by three parameters: (i) the number of clusters $n$; (ii) the cluster size $M$ and (iii) the number of days $T$, for which the study lasts. We consider a base simulation setting with $n=5$, $M=10$ and $T=14$ ($2$ weeks). We next vary each parameter independently to explore their effects: $n \in \{5, 10, 15, 20, 25, 30\}$, $M \in \{10, 20, 30, 40, 50\}$, and $T \in \{7, 14, 21, 28, 35\}$, while keeping the other two parameters fixed. The decision of actions is made on each day of a week. We also vary (iv) an additional parameter $\psi \in \{1,3,5,7,9\}$ which determines the intra-cluster correlations. This yields a total of $16\times 5=80$ simulation settings. For each setting, we repeats the experiments $20$ times with different seeds.

Comparison is made among the following algorithms/policies:
\begin{multicols}{2}
\begin{itemize}[leftmargin=*]
    \item \textbf{FQI} : The standard FQI (Algorithm \ref{alg:fqi}).
    \item \textbf{AGTD}: The adapted GTD (Algorithm 2).
    \item \textbf{GFQI}: The proposed GFQI with 
    exchangeable working correlation structure (Algorithm \ref{alg:optimal_fqi}).
    \item \textbf{CQL} \citep{kumar2020conservativeqlearningofflinereinforcement}: The conservative Q-learning algorithm that adds a regularization term to the loss function to 
    prevent the learned Q-values from being overestimated. 
    \item \textbf{DDQN} \citep{DBLP:journals/corr/HasseltGS15}: 
    An offline version of the double deep Q-network algorithm, designed to reduce the maximization bias by using two separate deep neural network models for the optimal Q-function, one for greedy action selection and the other for Q-value evaluation.
    \item \textbf{Behavior}: The uniform random policy used in the IHS study to collect the offline data.
\end{itemize}
\end{multicols}
Notice that the first five algorithms are all Q-learning-based. In particular, the first three algorithms employ linear polynomial basis functions to parameterize the optimal Q-function, with its degree selected from $\{1,2,3,4\}$ using 5-fold cross-validation. For GFQI and AGTD, the conditional expectation in \eqref{eqn:Phi*} 
at each iteration is estimated via linear regression. {\color{black}In practice, we recommend using supervised learning algorithms that are computationally efficient, as the proposed GFQI iteratively updates the Q-function, and each iteration requires to compute the conditional expectation in \eqref{eqn:Phi*}. More generally, kernel ridge regression can also be employed, as the kernel matrix only needs to be computed once and reused across iterations.} 
The covariance matrix $\bfV$ 
is updated at each iteration by imposing the exchangeable correlation structure among TD errors. The next two algorithms, CQL and DDQN, use deep neural networks for Q-function estimation.

Estimated optimal policies are evaluated via the Monte Carlo method. Specifically, we generate $10$ clusters' data that lasts for $35$ days, where each cluster contains $100$ trajectories. To minimize the evaluation error, we omit the reward residuals when generating the rewards. Finally, we  average these generated rewards across all trajectories to evaluate the expected return. 

\begin{figure}[t]
    \centering
    \includegraphics[width=0.60\linewidth]{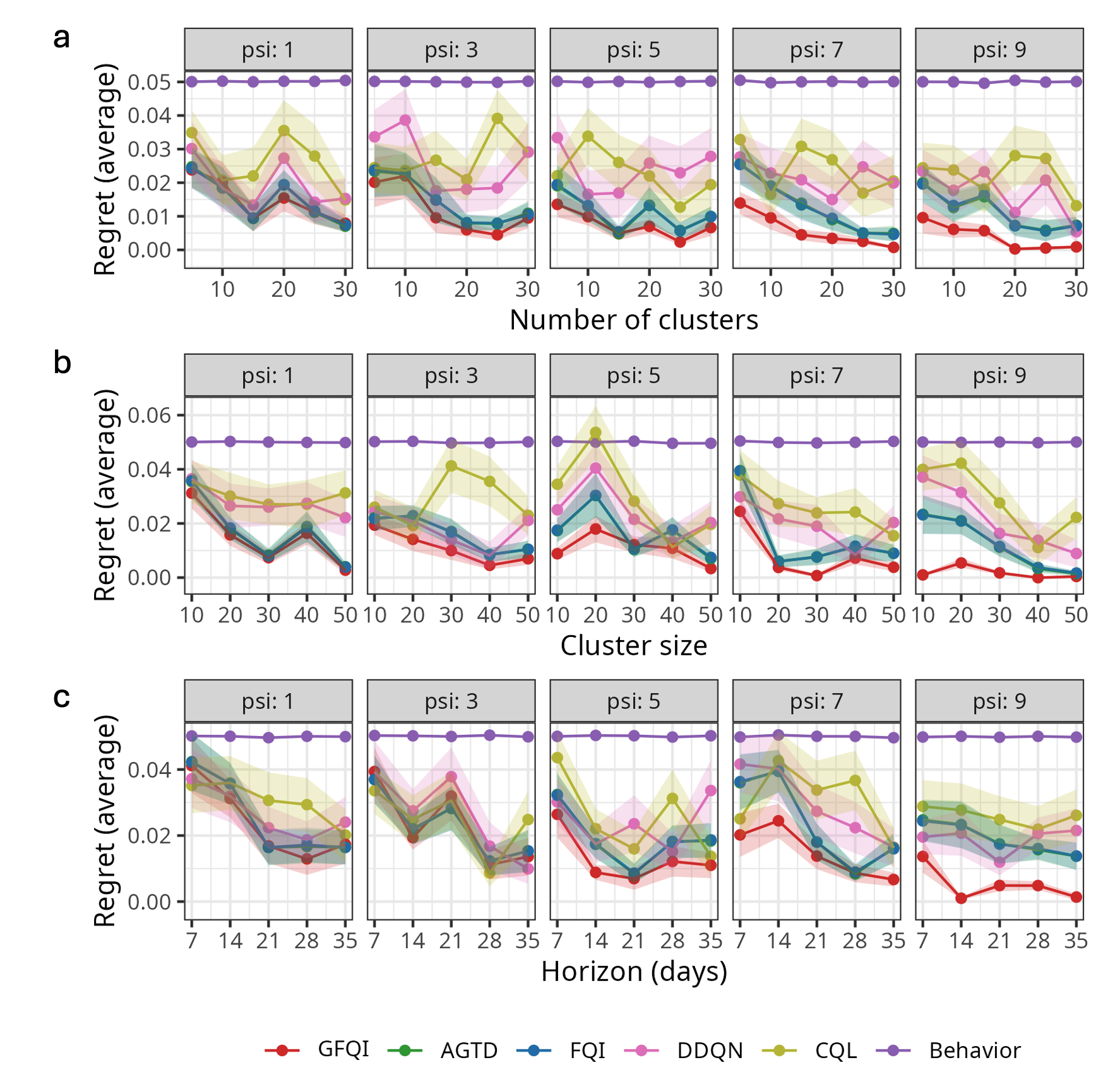}
    \caption{Regret of the average reward with varying (i) numbers of clusters, (ii) cluster sizes, (iii) numbers of days and (iv) values of intra-cluster correlation parameter ($\psi$). The shaded band represents the standard error. The green line (AGTD) and blue line (FQI) are largely overlapped. 
    }
    \label{fig:semi_all}
\end{figure}

Figure~\ref{fig:semi_all} visualizes the regrets of estimated policies computed by various RL algorithms with varying cluster sizes, numbers of clusters and horizons. We summarize our results as follows:
\begin{enumerate}[leftmargin=*]
    \item First, we find that GFQI considerably outperforms FQI and AGTD in scenarios with large intra-cluster correlations (i.e., $\psi\geq 5$) and achieves better performance in general with small correlations (i.e., $\psi\leq 3$). This result highlights the importance of accounting for the correlations in clustered data for policy learning.
    \item Second, the proposed GFQI consistently achieves smaller or comparable regrets than CQL and DDQN. This comparison implies that the danger of ignoring data correlations in policy learning can overshadow the benefits provided by the flexibility of deep neural network used in CQL and DDQN.
    \item Finally, all RL algorithms achieve significantly smaller regrets than the behavior policy in most cases, highlighting the value of policy learning in improving the medical outcomes of interns. 
\end{enumerate}

\section*{Acknowledgement}
Shi's research is partially supported by an EPSRC grant EP/W014971/1. 
The authors are grateful for the contributions of the researchers, administrators and participants involved in the Intern Health Study (https://clinicaltrials.gov/study/NCT03972293). The authors also thank the Editor, Associate Editor and reviewers for their comments, which have led to substantial improvements of the manuscript. 

\section*{Data Availability Statement}

Data are available from the authors with the permission of the PI of IHS, Dr. Srijan Sen.

\baselineskip=21pt
\bibliographystyle{apalike}
\bibliography{ref}

\begin{thebibliography}{}

\bibitem[Agarwal et~al., 2019]{agarwal2019reinforcement}
Agarwal, A., Jiang, N., Kakade, S.~M., and Sun, W. (2019).
\newblock Reinforcement learning: Theory and algorithms.
\newblock {\em CS Dept., UW Seattle, Seattle, WA, USA, Tech. Rep}, 32:96.

\bibitem[Andersen and Zhao, 2025]{andersen2025graphical}
Andersen, J.~B. and Zhao, Q. (2025).
\newblock A graphical approach to state variable selection in off-policy
  learning.
\newblock {\em arXiv preprint arXiv:2501.00854}.

\bibitem[Bian et~al., 2025]{bian2023off}
Bian, Z., Shi, C., Qi, Z., and Wang, L. (2025).
\newblock Off-policy evaluation in doubly inhomogeneous environments.
\newblock {\em Journal of the American Statistical Association},
  120(550):1102--1114.

\bibitem[Bradley, 2005]{bradley2005basic}
Bradley, R.~C. (2005).
\newblock Basic properties of strong mixing conditions. a survey and some open
  questions.
\newblock {\em Probability surveys}, 2:107--144.

\bibitem[Cai et~al., 2017]{cai2017real}
Cai, H., Ren, K., Zhang, W., Malialis, K., Wang, J., Yu, Y., and Guo, D.
  (2017).
\newblock Real-time bidding by reinforcement learning in display advertising.
\newblock In {\em Proceedings of the tenth ACM international conference on web
  search and data mining}, pages 661--670.

\bibitem[Carta et~al., 2021]{carta2021multi}
Carta, S., Corriga, A., Ferreira, A., Podda, A.~S., and Recupero, D.~R. (2021).
\newblock A multi-layer and multi-ensemble stock trader using deep learning and
  deep reinforcement learning.
\newblock {\em Applied Intelligence}, 51:889--905.

\bibitem[Chakraborty and Moodie, 2013]{chakraborty2013statistical}
Chakraborty, B. and Moodie, E.~E. (2013).
\newblock Statistical methods for dynamic treatment regimes.
\newblock {\em Springer-Verlag. doi}, 10(978-1):4--1.

\bibitem[Chen et~al., 2024]{chen2024reinforcement}
Chen, E.~Y., Song, R., and Jordan, M.~I. (2024).
\newblock Reinforcement learning in latent heterogeneous environments.
\newblock {\em Journal of the American Statistical Association}, accepted.

\bibitem[Chen and Jiang, 2019]{pmlr-v97-chen19e}
Chen, J. and Jiang, N. (2019).
\newblock Information-theoretic considerations in batch reinforcement learning.
\newblock In Chaudhuri, K. and Salakhutdinov, R., editors, {\em Proceedings of
  the 36th International Conference on Machine Learning}, volume~97 of {\em
  Proceedings of Machine Learning Research}, pages 1042--1051. PMLR.

\bibitem[Chen et~al., 2023]{chen2023steel}
Chen, X., Qi, Z., and Wan, R. (2023).
\newblock Steel: Singularity-aware reinforcement learning.
\newblock {\em arXiv preprint arXiv:2301.13152}.

\bibitem[Copas and Seaman, 2010]{doi:10.1080/02664760902939604}
Copas, A.~J. and Seaman, S.~R. (2010).
\newblock Bias from the use of generalized estimating equations to analyze
  incomplete longitudinal binary data.
\newblock {\em Journal of Applied Statistics}, 37(6):911--922.

\bibitem[Dabney et~al., 2018]{dabney2018distributional}
Dabney, W., Rowland, M., Bellemare, M., and Munos, R. (2018).
\newblock Distributional reinforcement learning with quantile regression.
\newblock In {\em Proceedings of the AAAI conference on artificial
  intelligence}, volume~32.

\bibitem[Dai et~al., 2020]{dai2020coindice}
Dai, B., Nachum, O., Chow, Y., Li, L., Szepesv{\'a}ri, C., and Schuurmans, D.
  (2020).
\newblock Coindice: Off-policy confidence interval estimation.
\newblock {\em Advances in neural information processing systems},
  33:9398--9411.

\bibitem[Diggle et~al., 2007]{10.1111/j.1467-9876.2007.00590.x}
Diggle, P., Farewell, D., and Henderson, R. (2007).
\newblock {Analysis of Longitudinal Data with Drop-Out: Objectives, Assumptions
  and a Proposal}.
\newblock {\em Journal of the Royal Statistical Society Series C: Applied
  Statistics}, 56(5):499--550.

\bibitem[Diggle et~al., 2002]{10.1093/oso/9780198524847.001.0001}
Diggle, P.~J., Heagerty, P.~J., Liang, K.-y., and Zeger, S.~L. (2002).
\newblock {\em {Analysis of Longitudinal Data}}.
\newblock Oxford University Press.

\bibitem[Duan et~al., 2024]{duan2024optimal}
Duan, Y., Wang, M., and Wainwright, M.~J. (2024).
\newblock Optimal policy evaluation using kernel-based temporal difference
  methods.
\newblock {\em The Annals of Statistics}, 52(5):1927--1952.

\bibitem[Ernst et~al., 2005]{JMLR:v6:ernst05a}
Ernst, D., Geurts, P., and Wehenkel, L. (2005).
\newblock Tree-based batch mode reinforcement learning.
\newblock {\em Journal of Machine Learning Research}, 6(18):503--556.

\bibitem[Ertefaie and Strawderman, 2018]{ertefaie2018constructing}
Ertefaie, A. and Strawderman, R.~L. (2018).
\newblock Constructing dynamic treatment regimes over indefinite time horizons.
\newblock {\em Biometrika}, 105(4):963--977.

\bibitem[Fan et~al., 2020]{pmlr-v120-yang20a}
Fan, J., Wang, Z., Xie, Y., and Yang, Z. (2020).
\newblock A theoretical analysis of deep q-learning.
\newblock In Bayen, A.~M., Jadbabaie, A., Pappas, G., Parrilo, P.~A., Recht,
  B., Tomlin, C., and Zeilinger, M., editors, {\em Proceedings of the 2nd
  Conference on Learning for Dynamics and Control}, volume 120 of {\em
  Proceedings of Machine Learning Research}, pages 486--489. PMLR.

\bibitem[Friedel et~al., 2019]{Friedel2019AMC}
Friedel, J.~E., DeHart, W.~B., Foreman, A.~M., and Andrew, M.~E. (2019).
\newblock A monte carlo method for comparing generalized estimating equations
  to conventional statistical techniques for discounting data.
\newblock {\em Journal of the experimental analysis of behavior}, 111
  2:207--224.

\bibitem[Garrett~Fitzmaurice and Molenberghs, 2008]{Garrett2008longitudinal}
Garrett~Fitzmaurice, Marie~Davidian, G.~V. and Molenberghs, G. (2008).
\newblock {\em Longitudinal Data Analysis}.
\newblock Chapman and Hall/CRC.

\bibitem[Hedeker~Donald, 2006]{Hedeker2006Longitudinal}
Hedeker~Donald, G. R.~D. (2006).
\newblock {\em Analysis of Longitudinal Data}.
\newblock John Wiley \& Sons.

\bibitem[Hu et~al., 2022]{hu2022doubly}
Hu, L., Li, M., Shi, C., Wu, Z., and Fryzlewicz, P. (2022).
\newblock Doubly inhomogeneous reinforcement learning.
\newblock {\em arXiv preprint arXiv:2211.03983}.

\bibitem[Hu and Wager, 2023]{hu2023off}
Hu, Y. and Wager, S. (2023).
\newblock Off-policy evaluation in partially observed markov decision processes
  under sequential ignorability.
\newblock {\em The Annals of Statistics}, 51(4):1561--1585.

\bibitem[Jin et~al., 2018]{Jin2018RealTimeBW}
Jin, J., Song, C.-N., Li, H., Gai, K., Wang, J., and Zhang, W. (2018).
\newblock Real-time bidding with multi-agent reinforcement learning in display
  advertising.
\newblock {\em Proceedings of the 27th ACM International Conference on
  Information and Knowledge Management}.

\bibitem[Jin et~al., 2021]{jin2021pessimism}
Jin, Y., Yang, Z., and Wang, Z. (2021).
\newblock Is pessimism provably efficient for offline rl?
\newblock In {\em International Conference on Machine Learning}, pages
  5084--5096. PMLR.

\bibitem[Kallus and Uehara, 2024]{kallus2024efficient}
Kallus, N. and Uehara, M. (2024).
\newblock Efficient evaluation of natural stochastic policies in off-line
  reinforcement learning.
\newblock {\em Biometrika}, 111(1):51--69.

\bibitem[Konda and Tsitsiklis, 1999]{konda1999actor}
Konda, V. and Tsitsiklis, J. (1999).
\newblock Actor-critic algorithms.
\newblock {\em Advances in neural information processing systems}, 12.

\bibitem[Kosorok and Laber, 2019]{kosorok2019precision}
Kosorok, M.~R. and Laber, E.~B. (2019).
\newblock Precision medicine.
\newblock {\em Annual review of statistics and its application}, 6(1):263--286.

\bibitem[Kosorok and Moodie, 2015]{kosorok2015adaptive}
Kosorok, M.~R. and Moodie, E.~E. (2015).
\newblock {\em Adaptive treatment strategies in practice: planning trials and
  analyzing data for personalized medicine}.
\newblock SIAM.

\bibitem[Krishnamurthy, 2016]{krishnamurthy2016partially}
Krishnamurthy, V. (2016).
\newblock {\em Partially observed Markov decision processes}.
\newblock Cambridge university press.

\bibitem[Kumar et~al.,
  2020]{kumar2020conservativeqlearningofflinereinforcement}
Kumar, A., Zhou, A., Tucker, G., and Levine, S. (2020).
\newblock Conservative q-learning for offline reinforcement learning.

\bibitem[Levine et~al., 2020]{levine2020offline}
Levine, S., Kumar, A., Tucker, G., and Fu, J. (2020).
\newblock Offline reinforcement learning: Tutorial, review, and perspectives on
  open problems.
\newblock {\em arXiv preprint arXiv:2005.01643}.

\bibitem[Li et~al., 2024]{li2024settling}
Li, G., Shi, L., Chen, Y., Chi, Y., and Wei, Y. (2024).
\newblock Settling the sample complexity of model-based offline reinforcement
  learning.
\newblock {\em The Annals of Statistics}, 52(1):233--260.

\bibitem[Li et~al., 2022]{li2022testing}
Li, M., Shi, C., Wu, Z., and Fryzlewicz, P. (2022).
\newblock Testing stationarity and change point detection in reinforcement
  learning.

\bibitem[Li et~al., 2025a]{Li2025supplement}
Li, M., Shi, C., Wu, Z., and Fryzlewicz, P. (2025a).
\newblock Supplement to ``testing stationarity and change point detection in
  reinforcement learning''.

\bibitem[Li et~al., 2025b]{li2025testing}
Li, M., Shi, C., Wu, Z., and Fryzlewicz, P. (2025b).
\newblock Testing stationarity and change point detection in reinforcement
  learning.
\newblock {\em The Annals of Statistics}, 53(3):1230--1256.

\bibitem[Li et~al., 2023]{li2023optimal}
Li, Z., Chen, J., Laber, E., Liu, F., and Baumgartner, R. (2023).
\newblock Optimal treatment regimes: a review and empirical comparison.
\newblock {\em International Statistical Review}, 91(3):427--463.

\bibitem[Liang and Zeger, 1986]{10.1093/biomet/73.1.13}
Liang, K.-y. and Zeger, S.~L. (1986).
\newblock {Longitudinal data analysis using generalized linear models}.
\newblock {\em Biometrika}, 73(1):13--22.

\bibitem[Liang and Recht, 2025]{liang2025randomization}
Liang, T. and Recht, B. (2025).
\newblock Randomization inference when n equals one.
\newblock {\em Biometrika}, 112(2):asaf013.

\bibitem[Liao et~al., 2021]{liao2021off}
Liao, P., Klasnja, P., and Murphy, S. (2021).
\newblock Off-policy estimation of long-term average outcomes with applications
  to mobile health.
\newblock {\em Journal of the American Statistical Association},
  116(533):382--391.

\bibitem[Liao et~al., 2022]{liao2022batch}
Liao, P., Qi, Z., Wan, R., Klasnja, P., and Murphy, S.~A. (2022).
\newblock Batch policy learning in average reward markov decision processes.
\newblock {\em Annals of statistics}, 50(6):3364.

\bibitem[Lipsitz et~al., 1994]{https://doi.org/10.1002/sim.4780131106}
Lipsitz, S.~R., Kim, K., and Zhao, L. (1994).
\newblock Analysis of repeated categorical data using generalized estimating
  equations.
\newblock {\em Statistics in Medicine}, 13(11):1149--1163.

\bibitem[Liu et~al., 2009]{Liu2009BivariateAA}
Liu, J., Pei, Y.-F., Papasian, C.~J., and Deng, H.-W. (2009).
\newblock Bivariate association analyses for the mixture of continuous and
  binary traits with the use of extended generalized estimating equations.
\newblock {\em Genetic Epidemiology}, 33.

\bibitem[Liu et~al., 2023]{liu2023online}
Liu, W., Tu, J., Zhang, Y., and Chen, X. (2023).
\newblock Online estimation and inference for robust policy evaluation in
  reinforcement learning.
\newblock {\em arXiv preprint arXiv:2310.02581}.

\bibitem[Liu et~al., 2020]{Liu2020FinRLAD}
Liu, X.-Y., Yang, H., Chen, Q., Zhang, R., Yang, L., Xiao, B., and Wang, C.
  (2020).
\newblock Finrl: A deep reinforcement learning library for automated stock
  trading in quantitative finance.
\newblock {\em Capital Markets: Market Microstructure eJournal}.

\bibitem[Luckett et~al., 2020]{luckett2020estimating}
Luckett, D.~J., Laber, E.~B., Kahkoska, A.~R., Maahs, D.~M., Mayer-Davis, E.,
  and Kosorok, M.~R. (2020).
\newblock Estimating dynamic treatment regimes in mobile health using
  v-learning.
\newblock {\em Journal of the american statistical association}.

\bibitem[Luo et~al., 2024]{luo2024policy}
Luo, S., Yang, Y., Shi, C., Yao, F., Ye, J., and Zhu, H. (2024).
\newblock Policy evaluation for temporal and/or spatial dependent experiments.
\newblock {\em Journal of the Royal Statistical Society Series B: Statistical
  Methodology}, page qkad136.

\bibitem[Ma et~al., 2023]{ma2023sequential}
Ma, T., Zhu, J., Cai, H., Qi, Z., Chen, Y., Shi, C., and Laber, E.~B. (2023).
\newblock Sequential knockoffs for variable selection in reinforcement
  learning.
\newblock {\em arXiv preprint arXiv:2303.14281}.

\bibitem[McCullagh, 2019]{mccullagh2019generalized}
McCullagh, P. (2019).
\newblock {\em Generalized linear models}.
\newblock Routledge.

\bibitem[McCulloch and Searle, 2000]{mcculloch2000generalized}
McCulloch, C.~E. and Searle, S.~R. (2000).
\newblock {\em Generalized, Linear, and Mixed Models}.
\newblock Wiley Series in Probability and Statistics. John Wiley \& Sons, Inc.
\newblock First published: 18 December 2000.

\bibitem[Miao et~al., 2025]{miao2025reinforcement}
Miao, R., Shahbaba, B., and Qu, A. (2025).
\newblock Reinforcement learning for individual optimal policy from
  heterogeneous data.
\newblock {\em arXiv preprint arXiv:2505.09496}.

\bibitem[Min et~al., 2021]{NEURIPS2021_3e6260b8}
Min, Y., Wang, T., Zhou, D., and Gu, Q. (2021).
\newblock Variance-aware off-policy evaluation with linear function
  approximation.
\newblock In Ranzato, M., Beygelzimer, A., Dauphin, Y., Liang, P., and Vaughan,
  J.~W., editors, {\em Advances in Neural Information Processing Systems},
  volume~34, pages 7598--7610. Curran Associates, Inc.

\bibitem[Mnih et~al., 2015]{MnihEtAl2015}
Mnih, V., Kavukcuoglu, K., Silver, D., et~al. (2015).
\newblock Human-level control through deep reinforcement learning.
\newblock {\em Nature}, 518:529--533.

\bibitem[Munos and Szepesv{{\'a}}ri, 2008]{JMLR:v9:munos08a}
Munos, R. and Szepesv{{\'a}}ri, C. (2008).
\newblock Finite-time bounds for fitted value iteration.
\newblock {\em Journal of Machine Learning Research}, 9(27):815--857.

\bibitem[NeCamp et~al., 2020]{necamp2020assessing}
NeCamp, T., Sen, S., Frank, E., Walton, M.~A., Ionides, E.~L., Fang, Y.,
  Tewari, A., and Wu, Z. (2020).
\newblock Assessing real-time moderation for developing adaptive mobile health
  interventions for medical interns: micro-randomized trial.
\newblock {\em Journal of medical Internet research}, 22(3):e15033.

\bibitem[Ouyang et~al., 2022]{ouyang2022training}
Ouyang, L., Wu, J., Jiang, X., Almeida, D., Wainwright, C., Mishkin, P., Zhang,
  C., Agarwal, S., Slama, K., Ray, A., et~al. (2022).
\newblock Training language models to follow instructions with human feedback.
\newblock {\em Advances in neural information processing systems},
  35:27730--27744.

\bibitem[Overall and Tonidandel, 2004]{2004Robustness}
Overall, J.~E. and Tonidandel, S. (2004).
\newblock Robustness of generalized estimating equation (gee) tests of
  significance against misspecification of the error structure model.
\newblock {\em Biometrical Journal}, 46(2).

\bibitem[Paradis and Claude, 2002]{PARADIS2002175}
Paradis, E. and Claude, J. (2002).
\newblock Analysis of comparative data using generalized estimating equations.
\newblock {\em Journal of Theoretical Biology}, 218(2):175--185.

\bibitem[Perdomo et~al., 2022]{perdomo2022complete}
Perdomo, J.~C., Krishnamurthy, A., Bartlett, P., and Kakade, S. (2022).
\newblock A complete characterization of linear estimators for offline policy
  evaluation.

\bibitem[Perdomo et~al., 2023]{perdomo2023complete}
Perdomo, J.~C., Krishnamurthy, A., Bartlett, P., and Kakade, S. (2023).
\newblock A complete characterization of linear estimators for offline policy
  evaluation.
\newblock {\em Journal of machine learning research}, 24(284):1--50.

\bibitem[Puterman, 2014]{puterman2014markov}
Puterman, M.~L. (2014).
\newblock {\em Markov decision processes: discrete stochastic dynamic
  programming}.
\newblock John Wiley \& Sons.

\bibitem[Qi et~al., 2025]{qi2025distributional}
Qi, Z., Bai, C., Wang, Z., and Wang, L. (2025).
\newblock Distributional off-policy evaluation in reinforcement learning.
\newblock {\em Journal of the American Statistical Association},
  (just-accepted):1--24.

\bibitem[Qin et~al., 2025]{qin2022reinforcement}
Qin, Z., Tang, X., Li, Q., Zhu, H., and Ye, J. (2025).
\newblock {\em Reinforcement Learning in the Ridesharing Marketplace}.
\newblock Springer.

\bibitem[Ramprasad et~al., 2023]{ramprasad2023online}
Ramprasad, P., Li, Y., Yang, Z., Wang, Z., Sun, W.~W., and Cheng, G. (2023).
\newblock Online bootstrap inference for policy evaluation in reinforcement
  learning.
\newblock {\em Journal of the American Statistical Association},
  118(544):2901--2914.

\bibitem[Riedmiller, 2005]{riedmiller2005neural}
Riedmiller, M. (2005).
\newblock Neural fitted q iteration--first experiences with a data efficient
  neural reinforcement learning method.
\newblock In {\em Machine learning: ECML 2005: 16th European conference on
  machine learning, Porto, Portugal, October 3-7, 2005. proceedings 16}, pages
  317--328. Springer.

\bibitem[Roderick~Little, 2019]{Roderick2019Statistical}
Roderick~Little, D.~R. (2019).
\newblock {\em Statistical analysis with missing data}.
\newblock Wiley.

\bibitem[Rubin, 1976]{10.1093/biomet/63.3.581}
Rubin, D.~B. (1976).
\newblock {Inference and missing data}.
\newblock {\em Biometrika}, 63(3):581--592.

\bibitem[Schulman et~al., 2015]{pmlr-v37-schulman15}
Schulman, J., Levine, S., Abbeel, P., Jordan, M., and Moritz, P. (2015).
\newblock Trust region policy optimization.
\newblock In Bach, F. and Blei, D., editors, {\em Proceedings of the 32nd
  International Conference on Machine Learning}, volume~37 of {\em Proceedings
  of Machine Learning Research}, pages 1889--1897, Lille, France. PMLR.

\bibitem[Shao et~al., 2024]{shao2024deepseekmath}
Shao, Z., Wang, P., Zhu, Q., Xu, R., Song, J., Bi, X., Zhang, H., Zhang, M.,
  Li, Y., Wu, Y., et~al. (2024).
\newblock Deepseekmath: Pushing the limits of mathematical reasoning in open
  language models.
\newblock {\em arXiv preprint arXiv:2402.03300}.

\bibitem[Shen and Chen, 2012]{10.1111/j.1541-0420.2012.01758.x}
Shen, C.-W. and Chen, Y.-H. (2012).
\newblock {Model Selection for Generalized Estimating Equations Accommodating
  Dropout Missingness}.
\newblock {\em Biometrics}, 68(4):1046--1054.

\bibitem[Shen and Chen, 2013]{Shen2013ModelSO}
Shen, C.-W. and Chen, Y.-H. (2013).
\newblock Model selection of generalized estimating equations with multiply
  imputed longitudinal data.
\newblock {\em Biometrical Journal}, 55.

\bibitem[Shen et~al., 2025]{shen2025deep}
Shen, G., Dai, R., Wu, G., Luo, S., Shi, C., and Zhu, H. (2025).
\newblock Deep distributional learning with non-crossing quantile network.
\newblock {\em arXiv preprint arXiv:2504.08215}.

\bibitem[Shi et~al., 2024a]{shi2024statistically}
Shi, C., Luo, S., Le, Y., Zhu, H., and Song, R. (2024a).
\newblock Statistically efficient advantage learning for offline reinforcement
  learning in infinite horizons.
\newblock {\em Journal of the American Statistical Association},
  119(545):232--245.

\bibitem[Shi et~al., 2024b]{shi2024value}
Shi, C., Qi, Z., Wang, J., and Zhou, F. (2024b).
\newblock Value enhancement of reinforcement learning via efficient and robust
  trust region optimization.
\newblock {\em Journal of the American Statistical Association},
  119(547):2011--2025.

\bibitem[Shi et~al., 2023a]{shi2023multiagent}
Shi, C., Wan, R., Song, G., Luo, S., Zhu, H., and Song, R. (2023a).
\newblock A multiagent reinforcement learning framework for off-policy
  evaluation in two-sided markets.
\newblock {\em The Annals of Applied Statistics}, 17(4):2701--2722.

\bibitem[Shi et~al., 2023b]{shi2023dynamic}
Shi, C., Wang, X., Luo, S., Zhu, H., Ye, J., and Song, R. (2023b).
\newblock Dynamic causal effects evaluation in a/b testing with a reinforcement
  learning framework.
\newblock {\em Journal of the American Statistical Association},
  118(543):2059--2071.

\bibitem[Shi et~al., 2022]{shi2022statistical}
Shi, C., Zhang, S., Lu, W., and Song, R. (2022).
\newblock Statistical inference of the value function for reinforcement
  learning in infinite-horizon settings.
\newblock {\em Journal of the Royal Statistical Society Series B: Statistical
  Methodology}, 84(3):765--793.

\bibitem[Shi et~al., 2024c]{shi2024off}
Shi, C., Zhu, J., Shen, Y., Luo, S., Zhu, H., and Song, R. (2024c).
\newblock Off-policy confidence interval estimation with confounded markov
  decision process.
\newblock {\em Journal of the American Statistical Association},
  119(545):273--284.

\bibitem[Silver et~al., 2016]{silver2016mastering}
Silver, D., Huang, A., Maddison, C.~J., Guez, A., Sifre, L., Van Den~Driessche,
  G., Schrittwieser, J., Antonoglou, I., Panneershelvam, V., Lanctot, M.,
  et~al. (2016).
\newblock Mastering the game of go with deep neural networks and tree search.
\newblock {\em nature}, 529(7587):484--489.

\bibitem[Sun et~al., 2024]{sun2024optimal}
Sun, K., Kong, L., Zhu, H., and Shi, C. (2024).
\newblock Arma-design: Optimal treatment allocation strategies for a/b testing
  in partially observable time series experiments.
\newblock {\em arXiv preprint arXiv:2408.05342}.

\bibitem[Sutton and Barto, 2018]{sutton2018reinforcement}
Sutton, R.~S. and Barto, A.~G. (2018).
\newblock {\em Reinforcement learning: An introduction}.
\newblock MIT press.

\bibitem[Tang et~al., 2019]{tang2019deep}
Tang, X., Qin, Z., Zhang, F., Wang, Z., Xu, Z., Ma, Y., Zhu, H., and Ye, J.
  (2019).
\newblock A deep value-network based approach for multi-driver order
  dispatching.
\newblock In {\em Proceedings of the 25th ACM SIGKDD international conference
  on knowledge discovery \& data mining}, pages 1780--1790.

\bibitem[Tsiatis et~al., 2019]{tsiatis2019dynamic}
Tsiatis, A.~A., Davidian, M., Holloway, S.~T., and Laber, E.~B. (2019).
\newblock {\em Dynamic treatment regimes: Statistical methods for precision
  medicine}.
\newblock CRC press.

\bibitem[Uehara et~al., 2022a]{uehara2022finitesampleanalysisminimax}
Uehara, M., Imaizumi, M., Jiang, N., Kallus, N., Sun, W., and Xie, T. (2022a).
\newblock Finite sample analysis of minimax offline reinforcement learning:
  Completeness, fast rates and first-order efficiency.

\bibitem[Uehara et~al., 2022b]{uehara2022review}
Uehara, M., Shi, C., and Kallus, N. (2022b).
\newblock A review of off-policy evaluation in reinforcement learning.
\newblock {\em arXiv preprint arXiv:2212.06355}.

\bibitem[Ueno et~al., 2011]{Ueno2011GeneralizedTL}
Ueno, T., ichi Maeda, S., Kawanabe, M., and Ishii, S. (2011).
\newblock Generalized td learning.
\newblock {\em J. Mach. Learn. Res.}, 12:1977--2020.

\bibitem[van Hasselt et~al., 2015]{DBLP:journals/corr/HasseltGS15}
van Hasselt, H., Guez, A., and Silver, D. (2015).
\newblock Deep reinforcement learning with double q-learning.
\newblock {\em CoRR}, abs/1509.06461.

\bibitem[Wang et~al., 2023]{wang2023projected}
Wang, J., Qi, Z., and Wong, R.~K. (2023).
\newblock Projected state-action balancing weights for offline reinforcement
  learning.
\newblock {\em The Annals of Statistics}, 51(4):1639--1665.

\bibitem[Wang et~al., 2025]{wang2025counterfactually}
Wang, J., Shi, C., Piette, J.~D., Loftus, J.~R., Zeng, D., and Wu, Z. (2025).
\newblock Counterfactually fair reinforcement learning via sequential data
  preprocessing.
\newblock {\em arXiv preprint arXiv:2501.06366}.

\bibitem[Wang et~al., 2011]{10.1111/j.1541-0420.2011.01678.x}
Wang, L., Zhou, J., and Qu, A. (2011).
\newblock {Penalized Generalized Estimating Equations for High-Dimensional
  Longitudinal Data Analysis}.
\newblock {\em Biometrics}, 68(2):353--360.

\bibitem[Wang et~al., 2024]{Wang2024GeneralizedEE}
Wang, Y.-W., Yang, H.-C., Chen, Y.-H., and Guo, C.-Y. (2024).
\newblock Generalized estimating equations boosting (geeb) machine for
  correlated data.
\newblock {\em Journal of Big Data}, 11:1--19.

\bibitem[Watkins and Dayan, 1992]{watkins1992q}
Watkins, C.~J. and Dayan, P. (1992).
\newblock Q-learning.
\newblock {\em Machine learning}, 8:279--292.

\bibitem[Wu et~al., 2020]{Wu2020AdaptiveST}
Wu, X., Chen, H., Wang, J., Troiano, L., Loia, V., and Fujita, H. (2020).
\newblock Adaptive stock trading strategies with deep reinforcement learning
  methods.
\newblock {\em Inf. Sci.}, 538:142--158.

\bibitem[Xu et~al., 2018]{xu2018large}
Xu, Z., Li, Z., Guan, Q., Zhang, D., Li, Q., Nan, J., Liu, C., Bian, W., and
  Ye, J. (2018).
\newblock Large-scale order dispatch in on-demand ride-hailing platforms: A
  learning and planning approach.
\newblock In {\em Proceedings of the 24th ACM SIGKDD international conference
  on knowledge discovery \& data mining}, pages 905--913.

\bibitem[Yang et~al., 2020]{Yang2020DeepRL}
Yang, H., Liu, X.-Y., Zhong, S., and Walid, A. (2020).
\newblock Deep reinforcement learning for automated stock trading: an ensemble
  strategy.
\newblock {\em Proceedings of the First ACM International Conference on AI in
  Finance}.

\bibitem[Yang et~al., 2022]{yang2022toward}
Yang, W., Zhang, L., and Zhang, Z. (2022).
\newblock Toward theoretical understandings of robust markov decision
  processes: Sample complexity and asymptotics.
\newblock {\em The Annals of Statistics}, 50(6):3223--3248.

\bibitem[Zeger and Liang, 1986]{ca1aab99-c8bf-3fed-8f2a-b3d28642c3ed}
Zeger, S.~L. and Liang, K.-Y. (1986).
\newblock Longitudinal data analysis for discrete and continuous outcomes.
\newblock {\em Biometrics}, 42(1):121--130.

\bibitem[Zhang et~al., 2023]{zhang2023estimation}
Zhang, L., Peng, Y., Liang, J., Yang, W., and Zhang, Z. (2023).
\newblock Estimation and inference in distributional reinforcement learning.
\newblock {\em arXiv preprint arXiv:2309.17262}.

\bibitem[Zhong et~al., 2020]{zhong2020risk}
Zhong, H., Deng, X., Fang, E.~X., Yang, Z., Wang, Z., and Li, R. (2020).
\newblock Risk-sensitive deep rl: Variance-constrained actor-critic provably
  finds globally optimal policy.
\newblock {\em arXiv preprint arXiv:2012.14098}.

\bibitem[Zhou et~al., 2023]{zhou2023distributional}
Zhou, W., Li, Y., Zhu, R., and Qu, A. (2023).
\newblock Distributional shift-aware off-policy interval estimation: A unified
  error quantification framework.
\newblock {\em arXiv preprint arXiv:2309.13278}.

\bibitem[Zhou et~al., 2024]{zhou2024estimating}
Zhou, W., Zhu, R., and Qu, A. (2024).
\newblock Estimating optimal infinite horizon dynamic treatment regimes via
  pt-learning.
\newblock {\em Journal of the American Statistical Association},
  119(545):625--638.

\bibitem[Zhu et~al., 2025]{zhu2025semi}
Zhu, J., Zhou, X., Yao, J., Aminian, G., Rivasplata, O., Little, S., Li, L.,
  and Shi, C. (2025).
\newblock Semi-pessimistic reinforcement learning.
\newblock {\em arXiv preprint arXiv:2505.19002}.

\bibitem[Zou et~al., 2019]{zou2019finite}
Zou, S., Xu, T., and Liang, Y. (2019).
\newblock Finite-sample analysis for sarsa with linear function approximation.
\newblock {\em Advances in neural information processing systems}, 32.

\end{thebibliography}

\appendix

This supporting information is organized as follows. Section \ref{sec:supp:proof} provides the technical assumptions required for our theoretical results and presents complete proofs of Theorems \ref{theo:asym properties} and \ref{theo:regret}. Section \ref{appendix:sec:simu} presents the data generating process for our semi-synthetic study (Section \ref{sec:numirical} in the main paper) along with additional simulation studies.

\section{Proofs}\label{sec:supp:proof}

\subsection{Assumptions}\label{sec:supp:ass}

We first impose the following assumptions.

\begin{asump}[Boundedness]\label{as:bounded reward}
    There exists some constant $0<R_{\max}<\infty$ such that $|R^{(j)}|\le R_{\max}$ almost surely for any $j$. 
\end{asump}
\begin{asump}[Realizability]\label{as:linear mdp}
    There exists some $\beta_0$ such that $\mathcal{R}(a,s)=\phi^\top(a,s)\beta_0$ for any $a$ and $s$. 
\end{asump}
\begin{asump}[Completeness]\label{as:completeness}
    For any $\bftheta$ whose $\ell_2$ norm $\|\bftheta\|_2\le 1$, there exists some $\bftheta^\prime$ with $\|\bftheta^\prime\|_2\leq 1$ such that the function $\mathcal{B} \phi^\top \bftheta=\phi^\top\bftheta^\prime$ where $\mathcal{B}$ denotes the operator $(\mathcal{B} g)(a,s) = \E[\max_{a^\prime} g(a^\prime, S')|A=a,S=s]$.
\end{asump}
\begin{asump}[Geometric ergodicity]\label{as:ergodic}
The state-action-reward $(\bfS,\bfA,\bfR)$ triplets over time forms a stationary and geometrically ergodic time series; see e.g., \citet{bradley2005basic} for the detailed definition.
\end{asump}
\begin{asump}[Stability]\label{as:bounded Dphi -gamma Dphi'}
    The matrix
    $$\frac{M^{-1}}{(1-\gamma)}
\E \left[\bm{\Phi}(\bfA,\bfS)\bm{\phi}^\top(\bfA,\bfS)-\gamma \bm{\Phi}(\bfA,\bfS)\bm{\phi}^\top(\pi^{*}(\bfS^\prime),\bfS^\prime)\right]
    $$ is invertible where $\bm{\phi}(\bfA,\bfS)$ is a shorthand for $[\phi(A^{(1)},S^{(1)}),\cdots, \phi(A^{(M)},S^{(M)})]$ and the expectation is taken with respect to the state-action pair's stationary distribution (see Assumption \ref{as:ergodic}). Additionally, the spectral norm of its inverse is bounded.
\end{asump}
\begin{asump}[Iteration number]\label{as:noofiterations}
    The number of GFQI iterations $K$ is much larger than $\log(N)/\log(\gamma^{-1})$. 
\end{asump}
\begin{asump}[Uniqueness]\label{as:unique optimal pi}
$\pi^{*}$ is unique. Additionally, the margin $\max_a Q^*(a,s)-\max_{a\neq \pi^*(s)} Q^*(a,s)$ is bounded away from zero.
\end{asump}
\begin{asump}[Value smoothness]\label{as:Q third derivatives}
    Let $\pi(\bftheta)$ be the greedy policy derived by $\phi(a,s)^\top\bftheta$. Then the expected cumulative reward for $\pi(\bftheta)$ has third-order derivative w.r.t $\bftheta$. 
\end{asump}
    
Next, we discuss the validity of the stability assumption (Assumption \ref{as:bounded Dphi -gamma Dphi'}). As commented in the main text, this assumption is likely to hold when the data was generated under the $\epsilon$-greedy algorithm. To elaborate, suppose the matrix $\bm{\Phi}$ takes the following form:
\begin{equation}\label{eqn:Phiform}
    \bm{\Phi}(\bfA,\bfS)=\underbrace{\Big[\phi(A^{(1)},S^{(1)}),\cdots, \phi(A^{(M)},S^{(M)})\Big]}_{\bm{\phi}(\bfA,\bfS)} \bm{\Omega},
\end{equation}
for some $M\times M$ positive-definite matrix $\bm{\Omega}$. Notice that the optimal $\bm{\Phi}^*$ satisfies this assumption by setting $\phi=\phi^*$ and $\bm{\Omega}=\bm{V}^{-1}$. Additionally, assume 
\begin{equation}\label{eqn:phiform}
\phi(A,S)=[\mathbb{I}(A=a_1)\phi^\top(S), \mathbb{I}(A=a_2)\phi^\top(S),\cdots,\mathbb{I}(A=a_{|\mathcal{A}|})\phi^\top(S)]^\top
\end{equation}
where $a_1,\cdots,a_{|\mathcal{A}|}$ denote the elements in the action space $\mathcal{A}$ and $\phi(S)$ denotes the state feature. 

To prove the invertibility of the matrix 
$$\bm{W}(\bm{\Phi})=\frac{1}{(1-\gamma)M}
\E \left[\bm{\Phi}(\bfA,\bfS)\bm{\phi}^\top(\bfA,\bfS)-\gamma \bm{\Phi}(\bfA,\bfS)\bm{\phi}^\top(\pi^{*}(\bfS^\prime),\bfS^\prime)\right],
$$it suffices to show that for any unit-norm vector $\bm{v}\in \mathbb{R}^d$, we have $\bm{\nu}^\top \bm{W}(\bm{\Phi}) \bm{\nu}\ge c$ for some constant $c>0$. This is because if $\bm{W}$ is singular, there must exist some $\bm{\nu}$ such that $\bm{W}(\bm{\Phi}) \bm{\nu}=0$ and hence $\bm{\nu}^\top \bm{W}(\bm{\Phi}) \bm{\nu}=0$. 

When $\bm{\Phi}$ satisfies \eqref{eqn:Phiform}, it suffices to show that 
\begin{equation}\label{eqn:somekeyinequality}
    \frac{1}{(1-\gamma)M}  \Big[\mathbb{E}\|\bm{\nu}^\top \widetilde{\bm{\phi}}(\bfA,\bfS)\|_2^2- \gamma \mathbb{E}|\bm{\nu}^\top \widetilde{\bm{\phi}}(\bfA,\bfS)\widetilde{\bm{\phi}}^\top(\pi^*(\bfS^\prime), \bfS^\prime) \bm{\nu}|\Big]\ge c,
\end{equation}
where $\widetilde{\bm{\phi}}(\bfA,\bfS)=\bm{\phi}(\bfA,\bfS)\bm{\Omega}^{1/2}$ and $\bm{\Omega}^{1/2}$ denotes the symmetric matrix such that $\bm{\Omega}^{1/2}\bm{\Omega}^{1/2}=\bm{\Omega}$. Since $\bm{\Omega}$ is positive-definite, its minimum eigenvalue is lower bounded by some constant $\bar{c}>0$. This together with the fact that $(S^{(1)},A^{(1)}),\cdots,(S^{(M)},A^{(M)})$ follow the same distribution, we obtain that
\begin{equation}\label{eqn:minimumeigenvalue}
    \frac{1}{M}\mathbb{E}\|\bm{\nu}^\top \widetilde{\bm{\phi}}(\bfA,\bfS)\|_2^2\ge \frac{\bar{c}}{M}\mathbb{E}\|\bm{\nu}^\top \bm{\phi}(\bfA,\bfS)\|_2^2=\bar{c}\mathbb{E}|\bm{\nu}^\top \phi(A,S)|^2.
\end{equation}
When $\phi$ satisfies \eqref{eqn:phiform} with proper choice of the state feature, the rightmost term in \eqref{eqn:minimumeigenvalue} is strictly positive; see e.g., Appendix C1 of \citet{shi2022statistical}. 

By Cauchy-Schwarz inequality, the second term on the left-hand-side (LHS) of \eqref{eqn:somekeyinequality} within the square brackets is smaller than or equal to $\gamma \sqrt{\mathbb{E}\|\bm{\nu}^\top \widetilde{\bm{\phi}}(\bfA,\bfS)\|_2^2}\sqrt{\mathbb{E}\|\widetilde{\bm{\phi}}^\top(\pi^*(\bfS^\prime), \bfS^\prime) \bm{\nu}\|_2^2}$. Due to that $M^{-1} \mathbb{E}\|\bm{\nu}^\top \widetilde{\bm{\phi}}(\bfA,\bfS)\|_2^2$ is bounded away from zero, it suffices to show that
\begin{equation*}
    \frac{1}{(1-\gamma)\sqrt{M}}\Big[\sqrt{\mathbb{E}\|\bm{\nu}^\top \widetilde{\bm{\phi}}(\bfA,\bfS)\|_2^2}- \gamma \sqrt{\mathbb{E}\|\widetilde{\bm{\phi}}^\top(\pi^*(\bfS^\prime), \bfS^\prime) \bm{\nu}\|_2^2}\Big]
\end{equation*}
is bounded away from zero, in order to prove \eqref{eqn:somekeyinequality}. Using similar arguments to proving the positiveness of $M^{-1} \mathbb{E}\|\bm{\nu}^\top \widetilde{\bm{\phi}}(\bfA,\bfS)\|_2^2$, we can show that both $M^{-1} \mathbb{E}\|\bm{\nu}^\top \widetilde{\bm{\phi}}(\bfA,\bfS)\|_2^2$ and $M^{-1} \mathbb{E}\|\widetilde{\bm{\phi}}^\top(\pi^*(\bfS^\prime), \bfS^\prime) \bm{\nu}\|_2^2$ are bounded away from infinity. As such, it suffices to show
\begin{eqnarray*}
    &&\frac{1}{(1-\gamma)M}\Big[\mathbb{E}\|\bm{\nu}^\top \widetilde{\bm{\phi}}(\bfA,\bfS)\|_2^2- \gamma^2 \mathbb{E}\|\widetilde{\bm{\phi}}^\top(\pi^*(\bfS^\prime), \bfS^\prime) \bm{\nu}\|_2^2\Big]\\
    &=&\frac{1}{(1-\gamma)\sqrt{M}}\Big[\sqrt{\mathbb{E}\|\bm{\nu}^\top \widetilde{\bm{\phi}}(\bfA,\bfS)\|_2^2}- \gamma \sqrt{\mathbb{E}\|\widetilde{\bm{\phi}}^\top(\pi^*(\bfS^\prime), \bfS^\prime) \bm{\nu}\|_2^2}\Big]\\
    &\times& \frac{1}{\sqrt{M}}\Big[\sqrt{\mathbb{E}\|\bm{\nu}^\top \widetilde{\bm{\phi}}(\bfA,\bfS)\|_2^2}+\gamma \sqrt{\mathbb{E}\|\widetilde{\bm{\phi}}^\top(\pi^*(\bfS^\prime), \bfS^\prime) \bm{\nu}\|_2^2}\Big],
\end{eqnarray*}
is bounded away from zero. Using similar arguments to the proof of \eqref{eqn:minimumeigenvalue}, it suffices to show 
\begin{eqnarray*}
    \frac{1}{1-\gamma} \mathbb{E}\Big[|\bm{\nu}^\top \phi(A,S)|^2-\gamma^2 |\bm{\nu}^\top \phi(\pi^*(S'),S')|^2\Big]
\end{eqnarray*}
is bounded away from zero, or equivalently,
\begin{eqnarray}\label{eqn:finalcon}
    \lambda_{\min}\Big[ \mathbb{E} \{\phi(A,S)\phi^\top(A,S)-\gamma^2 \phi(\pi^*(S'),S')\phi^\top(\pi^*(S'),S')\} \Big]\ge c^*,
\end{eqnarray}
for some constant $c^*>0$. Condition \eqref{eqn:finalcon} is automatically satisfied under certain suitable choices of the state feature vector, provided that the behavior policy is $\epsilon$-greedy with respect to the optimal policy $\pi^*$. Refer to Appendix C1 of \citet{shi2022statistical} for the proof. 

So far, we have proven the invertibility of $\bm{W}(\bm{\Phi})$. For any vector $\bm{\mu}$, we can construct a unit-norm vector $\bm{W}^{-1}(\bm{\Phi}) \bm{\mu}/\|\bm{W}^{-1}(\bm{\Phi}) \bm{\mu}\|_2$. By setting $\bm{\nu}$ to $\bm{W}^{-1}(\bm{\Phi}) \bm{\mu}/\|\bm{W}^{-1}(\bm{\Phi}) \bm{\mu}\|_2$, we obtain that
\begin{eqnarray*}
    \bm{\mu}^\top \bm{W}^{-1}(\bm{\Phi}) \bm{\mu}\ge c\|\bm{W}^{-1}(\bm{\Phi}) \bm{\mu}\|_2^2.
\end{eqnarray*}
By Cauchy-Schwarz inequality, the LHS is smaller than or equal to $\|\bm{\mu}\|_2 \|\bm{W}^{-1}(\bm{\Phi}) \bm{\mu}\|_2$. It follows that $\|\bm{W}^{-1}(\bm{\Phi}) \bm{\mu}\|_2/\|\bm{\mu}\|_2\le c^{-1}$. This proves the boundedness of the spectral norm of $\|\bm{W}^{-1}(\bm{\Phi})\|_2$. 

\subsection{Proof of Theorem \ref{theo:asym properties}}\label{sec:proof th1}



\textbf{Proof of the MSE of the estimator}.
To simplify notation, we focus on a finite MDP where both $\mathcal{S}$ and $\mathcal{A}$ are discrete. 
The proof is based on those for Lemmas B.1 -- B.3 in \citet{li2022testing}, which establish the rate of convergence and asymptotic normality of $\widehat{\beta}$ when employing the standard FQI with linear function approximation.  Below, we focus on outlining the key steps of our proof to save space. 

Notice that in GFQI, we iteratively update the
Q-function according to the formula,
$$
\sum_{(\bfS, \bfA,\bfR,\bfS^\prime)}
    \bm{\Phi}(\bfA,\bfS) 
    \bm{\delta}(\bfS,\bfA,\bfR,\bfS';\bftheta^{(k)}, \bftheta^{(k-1)})=0. 
$$
At the $k$th iteration, we define the population-level Q-function
\begin{equation*}
Q^{(k+1), *}(a, s)=\mathcal{R}(a, s)+\gamma  
\sum_{s^{\prime}}\max _{a^{\prime}} 
Q^{(k)}\left(a^{\prime}, s^{\prime}\right) \mathcal{T}\left(s^{\prime} \mid a, s\right).
\end{equation*}
According to the Bellman optimality equation,
$$
Q^{*}(a, s)=\mathcal{R}(a, s)+\gamma  \sum_{s^{\prime}}\max _{a^{\prime}} Q^{*}\left(a^{\prime}, s^{\prime}\right) \mathcal{T}\left(s^{\prime} \mid a, s\right) .
$$
It follows that
\begin{equation*}
\label{eq:optimal-k+1}
    \begin{split}
        &\sup _{a, s}\left|Q^{*}(a, s)-Q^{(k+1)}(a, s)\right| \\
&\leq  \sup _{a, s}\left|Q^{(k+1), *}(a, s)-Q^{(k+1)}(a, s)\right|+\sup _{a, s}\left|Q^{*}(a, s)-Q^{(k+1), *}(a, s)\right| \\
& \leq \sup _{a, s}\left|Q^{(k+1), *}(a, s)-Q^{(k+1)}(a, s)\right|+\gamma \sup _{a, s}\left|Q^{*}(a, s)-Q^{(k)}(a, s)\right| .
    \end{split}
\end{equation*}
Iteratively applying this inequality for $k=K, K-1, \cdots, 1$, we obtain that
\begin{equation*}
\sup _{a, s}\left|Q^{*}(a, s)-\widehat{Q}(a, s)\right|\leq \sum_k \gamma^{K-k} \sup _{a, s}\left|Q^{(k+1), *}(a, s)-Q^{(k+1)}(a, s)\right| +\gamma^{K+1} \sup _{a, s}\left|Q^{*}(a, s)-Q^{(0)}(a, s)\right|.
\end{equation*}
As $K$ diverges to infinity, the second term on the RHS decays to zero. The first term is upper bounded by
\begin{equation}\label{eq:FQI error first term}
\frac{1}{1-\gamma} \sup _{a, s, k}\left|Q^{(k+1), *}(a, s)-Q^{(k+1)}(a, s)\right| .
\end{equation}
When employing classical FQI, this term has been shown to decay to zero as the sample size grows to infinity, and the associated error bound has been established \citep[see, e.g.,][Lemma B.2]{li2022testing}. The key difference in our approach lies in the use of GEE to update the Q-function at each iteration. Nevertheless, under the GEE version of the stability condition outlined in Assumption \ref{as:bounded Dphi -gamma Dphi'}, the arguments used to prove Lemma B.2 of \citet{li2022testing} remain applicable. Consequently, we can show that \eqref{eq:FQI error first term} is of the order $O((1-\gamma)^{-2}\sqrt{N^{-1}\log N}R_{\max})$, with probability at least $1-O(N^{-\kappa})$ for any sufficiently large constant $\kappa>0$. 

Consequently, under the given conditions on $K$ in Assumption \eqref{as:noofiterations}, we obtain that
$$
\max _{K / 2\leq k \leq K} \sup _{a, s}\left|Q^{*}(a, s)-Q^{(k+1)}(a, s)\right|=O\Big(\frac{R_{\max}\sqrt{N^{-1}\log N}}{(1-\gamma)^2}\Big),
$$
with probability at least $1-O(N^{-\kappa})$. 

{\color{black}Since $R_{\max}$ is a constant (see Assumption \ref{as:bounded reward}), and the discount factor $\gamma$ is strictly smaller than $1$, the above uniform error bound converges to zero as $N$ approaches infinity. By Assumption \ref{as:unique optimal pi}, the margin $\max_a Q^*(a,s)-\max_{a\neq \pi^*(s)} Q^*(a,s)$ is bounded away from zero. Consequently, for some sufficiently large $N$, the error bound will be strictly smaller than the margin. As a result, $\arg \max _a \phi_L^{\top}(a, s) \beta^{(k)}$ will be exactly equal to $\pi^*$ for sufficiently large $N$.} As such, for any $k\ge K/2$, $\bftheta^{(k)}$ is the solution to the following estimating equations:
$$
\sum_{(\bfS, \bfA,\bfR,\bfS^\prime)}
    \bm{\Phi}(\bfA,\bfS) 
    \bm{\delta}(\bfS,\bfA,\bfR,\bfS';\bftheta)=0. 
$$
Recall that $\bm{W}(\bm{\Phi})$ denotes the matrix 
\begin{eqnarray*}
    \bm{W}(\bm{\Phi})=\frac{1}{(1-\gamma)M} \mathbb{E}\bm{\Phi}(\bfA,\bfS) [\bm{\phi}(\bm{A},\bm{S})-\gamma \bm{\phi}(\pi(\bm{\bfS}'), \bm{\bfS}')]^\top.
\end{eqnarray*}
Using similar arguments to the asymptotic analysis in the proof of Lemma B.2 of \citet{li2022testing}, we can obtain the following linear representation for $\widehat{\beta}$,  given by
\begin{align*}
    \widehat{\bftheta}-\bftheta^*=\underbrace{\frac{\bm{W}^{-1}(\bm{\Phi})}{(1-\gamma)MN}\sum_{(\bfS, \bfA,\bfR,\bfS^\prime)}
    \bm{\Phi}(\bfA,\bfS)
    \bm{\delta}(\bfS,\bfA,\bfR,\bfS';\bftheta^{*})}_{\textrm{leading~term}}+\underbrace{O\Big(\frac{R_{\max}\log(N)}{(1-\gamma)^3 N}\Big)}_{\textrm{reminder~term}},
\end{align*}
with probability at least $1-O(N^{-\kappa})$. Denote this event as $B$. The MSE can be decomposed as follows: 
\begin{eqnarray}\label{eqn:somesomeequation}
    \E \|\widehat{\bftheta} - \bftheta \|_2^2 = \E\|\widehat{\bftheta}-\bftheta\|_2^2 \mathbb{I}(B) + \E\|\widehat{\bftheta}-\bftheta\|_2^2 \mathbb{I}(B^c). 
\end{eqnarray}
The second term is of high-order given sufficiently large $N$, since the event $B^c$ (the complement of $B$) occurs with probability $N^{-\kappa}$ for some sufficiently large $\kappa$. 

It remains to upper bound the first term. Based on the linear representation, the first term consists of the three terms: (i) the trace of $\mathbb{E}\|\textrm{leading~term}\|_2^2$, which corresponds to the leading asymptotic variance term in Theorem \ref{theo:asym properties}; (ii) the trace of $\mathbb{E}\|(\textrm{leading~term})^\top (\textrm{reminder~term})\|_2^2$ which can be upper bounded by $\sqrt{\mathbb{E}\|\textrm{leading~term}\|_2^2\mathbb{E}\|\textrm{reminder~term}\|_2^2}$ according to Cauchy-Schwarz inequality, whose order of magnitude is given by the finite-sample bias term in Theorem \ref{theo:asym properties}; (iii) $\mathbb{E}\|\textrm{reminder~term}\|_2^2$ whose order of magnitude is smaller than (ii) for sufficiently large $N$. The proof is hence completed.

\smallskip

\noindent \textbf{Proof of efficiency}. We next prove that for any $\bm{\Phi}(\bfA,\bfS)$, the asymptotic variance of the solution to \eqref{eq:optimal bellman} satisfies the following inequality:
$$
\bm{W}^{-1}(\bm{\Phi})\bm{\Sigma}(\bm{\Phi})\bm{W}^{-1\top}(\bm{\Phi})\geq \bm{W}^{-1}(\bm{\Phi}^*)\bm{\Sigma}(\bm{\Phi}^*)\bm{W}^{-1\top}(\bm{\Phi}^*)=\frac{1}{1-\gamma}\bm{W}^{-1}(\bm{\Phi}^*).
$$
We remove the dependence on $\bfA$ and $\bfS$ for simplification in notation and denote $\bm{\Phi}(\bfA,\bfS)$, $\bm{\phi}(\bfA,\bfS)$, $\E(\bm{\phi}(\pi^*(\bfS^\prime),\bfS^\prime)|\bfA,\bfS)$ by $\bm{\Phi}$, $\bm{\phi}$ and $\bm{\phi}^\prime$, respectively.
It is equivalent to show
$$
\frac{1}{1-\gamma}\bm{a}^\top\bm{\Sigma}^{-1/2}(\bm{\Phi}) \bm{W}(\bm{\Phi}) \bm{W}^{-1}(\bm{\Phi}^*)(\bm{\Phi}^*)\bm{W}^\top(\bm{\Phi})\bm{\Sigma}^{-1/2\top}(\bm{\Phi})\bm{a}\leq \|\bm{a}\|_2^2,
$$
for any $\bm{a}\in \mathbb{R}^{d}$.
By Cauchy–Schwarz inequality, the LHS 
$$
\begin{aligned}
&\|\bm{a}^\top\E^{-1/2}(\bm{\Phi}\bfV\bm{\Phi}^{\top})
\E[\bm{\Phi}(\bm{\phi} - \gamma \bm{\phi}^\prime)] 
\E^{-1/2}[(\bm{\phi} - \gamma \bm{\phi}^\prime)\bfV^{-1}(\bm{\phi} - \gamma \bm{\phi}^\prime)^\top]
\|_2^2\\
\leq &\|\bm{a}^\top \E^{-1/2}\left(\bm{\Phi}\bfV \bm{\Phi}^{\top} \right)
\E^{1/2}(\bm{\Phi}\bfV\bm{\Phi})
\E^{1/2}[(\bm{\phi} - \gamma \bm{\phi}^\prime)\bfV^{-1}(\bm{\phi} - \gamma \bm{\phi}^\prime)^\top]
\E^{-1/2}[(\bm{\phi} - \gamma \bm{\phi}^\prime)\bfV^{-1}(\bm{\phi} - \gamma \bm{\phi}^\prime)^\top]
\|_2^2\\
=& \|\bm{a}\|_2^2.
\end{aligned}
$$
Therefore, we have shown that 
the solution to the following GEE
\begin{equation*}
\begin{aligned}
\mathbf{0}= \sum_{i,t} \left\{\phi(\Ait,\Sit) -\gamma \E\left(\phi(\pi^{*}(\Sit[i][t+1]), \Sit[i][t+1])\mid \Ait,\Sit\right)\right\}\Vi^{-1}\\\times \left\{\Rit + \gamma \phi(\pi^{*}(\Sit[i][t+1]), \Sit[i][t+1]) \bftheta - \phi(\Ait,\Sit) \bftheta \right\}.
\end{aligned}
\end{equation*}
 has the minimal asymptotic variance among the class
of estimators computed by solving \eqref{eq:optimal bellman}. This completes the proof.

\subsection{Proof of Theorem \ref{theo:regret}}\label{sec:proof th2}
Notice that regret can be represented by $\mathcal{V}(\beta^*)-\E \mathcal{V}(\widehat{\beta})$. Applying a Taylor expansion of $\mathcal{V}(\widehat{\beta})$ around $\beta^*$, the first-order term diminishes as $\beta^*$ corresponds to the argmax of $\mathcal{V}$. It then follows that
$$
\begin{aligned}
    \mathcal{V}(\beta^*)-\E \mathcal{V}(\widehat{\beta})
    = -\frac{1}{2}\E(\widehat{\bftheta}-\bftheta^*)^\top H(\widehat{\bftheta}- \bftheta^*)
    + O(\E \| \widehat{\bftheta} - \bftheta^*\|_2^3).
\end{aligned}
$$
Under the linear representation established in Theorem \ref{theo:asym properties}, using similar arguments in \eqref{eqn:somesomeequation}, it can be shown that the first term on the right-hand-side equals
\begin{eqnarray*}
    -\frac{1}{2}\mathrm{tr}(\Var_A(\widehat{\bftheta})H)+ 
    O\Big(\frac{R_{\max}\log(N)}{(1-\gamma)^{5} N^{\frac{3}{2}}}\Big). 
\end{eqnarray*}
As for the second term, it is of the order of magnitude $O((1-\gamma)^{-6}R_{\max}^{3/2}N^{-3/2}\log(N))$. The proof is hence completed. 

{\color{black}Finally, we remark that while the regret bound depends explicitly on $N$, the number of clusters, it does not depend on $M$, the number of subjects per cluster. This is because, although the total sample size increases with $M$, we do not impose any assumption on the within-cluster correlations. As a result, without further structural assumptions, the effective sample size is determined solely by $N$, and not by $M$.}

\section{Additional Details on Empirical Studies}
\label{appendix:sec:simu}
\subsection{More on the Semi-synthetic Study}
\textbf{More on Simulation Design}. 
Simulated data is generated as follows:
\begin{align*}
	S_t^{(i,j)} &= f\left(S_{t-1}^{(i,j)},A_{t-1}^{(i,j)}\right) + \varepsilon_{t}^{(i)} + \epsilon_t^{(i,j)} , \\
        R_t^{(i,j)} &= \mathcal{R}\left(S_{t-1}^{(i,j)},A_{t-1}^{(i,j)}\right) + \alpha_{t}^{(i)} + \xi_t^{(i,j)},
\end{align*}
where $i$ is the index for team, $j$ is the index for intern, and $t$ is the index for week. Mean models $f$ and $\mathcal{R}$ are trained from the IHS dataset using polynomial regression with degree of $2$. $\varepsilon_{t}^{(i)}$ and $\alpha_{t}^{(i)}$ denote team-specific random effects shared among all interns with the same team $i$ at week $t$, and $\epsilon_t^{(i,j)}$ and $\xi_t^{(i,j)}$ are intern-specific measurement errors, which are independent across different interns. 

All these errors are set to follow Gaussian distributions with means equal to zero and variances given by $\sigma_1^2$, $\sigma_2^2$, $\sigma_3^2$ and $\sigma_4^2$. More specifically,  
\begin{align*}
    \epsilon_{t}^{(i,j)} &\sim \mathcal{N}(0,\sigma_1^2), \varepsilon_{t}^{(i)} \sim \mathcal{N}(0,\sigma_2^2) \\
    \xi_{t}^{(i,j)} &\sim \mathcal{N}(0,\sigma_3^2), \alpha_{t}^{(i)} \sim \mathcal{N}(0,\sigma_4^2).
\end{align*}
Additionally, we introduce a new parameter $\psi$ to control the magnitude of correlation within each team. To keep the marginal variances of states and rewards to be invariant to $\psi$, we set $\{\sigma_i^2\}_{i=1,\dots,4}$ as below,

\begin{align*}
    \left[\sigma_{1}^2,\sigma_2^2\right]^\top &= \sigma_{s}^2 *\left[1 - \psi\rho_{s}^2,\psi\rho_{s}^2 \right]^\top, \\
    \left[\sigma_{3}^2,\sigma_4^2\right]^\top &= \sigma_{r}^2 *\left[1 - \psi\rho_{r}^2,\psi\rho_{r}^2 \right]^\top,
\end{align*}
where $\sigma_{s}^2$ and $\sigma_{r}^2$ are the marginal state and reward variances learned from the data, $\rho_{s}$ and $\rho_{r}$ are the within-cluster correlation coefficient for state and reward, which are determined through linear mixed model \citep{mcculloch2000generalized}. Hence, a larger $\psi$  increases the variances of the team-specific random effects $\varepsilon_t$ and $\alpha_t$. This, in turn, results in higher correlations among interns within the same team.

In this semi-synthetic data analysis, we set the daily cubic root of step count as the state variable, and daily mood score as the reward variable. The estimated $\hat{\sigma}_s^2 = 11.5$, $\hat{\rho}_s^2 = 0.07$, $\hat{\sigma}_r^2 = 2.2$, and $\hat{\rho}_r^2 = 0.09$.

The behavior policy that generates the simulated data is set to a uniform random policy with $\mathbb{P}(A_t = 1) = \mathbb{P}(A_t = 0) = 0.5$.

\noindent \textbf{Implementation Detail}. For CQL and DDQN, the Q function is approximated by a fully connected feed-forward neural network consisting of three sequential layers: the first with $1$ input and $16$ output features, the second with $16$ input and $16$ output features, and the third with $16$ input and $2$ output features, where the first two layers are followed by ReLU activation functions. The standard error in Figure \ref{fig:semi_all} is calculated with respect to the random seed after running experiments $20$ times. The simulation were conducted on Google Cloud using "n2-standard" machine type (CPU only).

\subsection{Simulation Study}\label{sec:simulation}
    
The primary objective of this simulation study is to evaluate the  policy regret under various correlation structures. The study examines the effect of different parameters such as the number of clusters, time horizon, and cluster size on the performance of the policies. 
The simulations were conducted
on the school's
high-performance computing cluster
using CPUs. 


\noindent \textbf{Correlation Design.}
The correlation structures were designed to reflect different potential dependencies within the data. The state update was modeled as:
$\Sijt[i][m][t+1] = 0.5\Sijt (2\Aijt-1)+\beta_{t}^{(i)}$
where $\beta_{t}^{(i)}\sim \mathcal{N}(0, \sigma_1^2)$ with $\sigma_1^2 =0.25$. 
The reward was expressed as:
$\Rijt=0.25(\Sijt)^2(2\Aijt-1)+\Sijt+\alpha_t^{(i)}+ \epsilon_t^{(i,m)}$, with $\epsilon_{t}^{(i,m)}\sim \mathcal{N}(0, \sigma_2^2)$ and $\alpha_t^{(t)}\sim \mathcal{N}(0, \sigma_3^2)$, where $\sigma_2^2 = 0.25$ and $\sigma_3^2=4$.
This correlation structure can be approximated by the exchangeable structure.

\noindent \textbf{Simulation Parameters.}
The simulation was conducted by varying three key parameters:
\begin{itemize}
    \item Number of Teams: 5, 10, 15, 20, 25, 30 (with cluster size = 5, horizon = 5)
    \item Time Horizon: 5, 10, 15, 20, 25, 30 (with number of clusters = 5, cluster size = 5)
    \item Team Size: 5, 10, 15, 20, 25, 30 (with number of clusters = 5, horizon = 5)
\end{itemize}

Each scenario was repeated 50 times to account for variability in the results.

\noindent \textbf{Policies Compared.}
Comparison is made among
the following algorithms/policies specified in Section \ref{sec:numirical}.
For CQL and DDQN, the Q function is approximated by a fully connected feed-forward neural network consisting of four sequential layers: the first with 1 input and 8 output features, the second and third with 8 input and 8 output features, and the forth with 8 input and 2 output features, where the first two layers are followed by ReLU activation functions. 

\noindent \textbf{Evaluation.}
Policies were evaluated by executing each to generate 100 trajectories of length 1000. The average of the average rewards and the cumulative discounted rewards across these trajectories were calculated.

We compare the average reward of the offline trained policies with a policy trained in an online, uncorrelated environment
with Q-learning approach utilizing a neural network  approximator with a 64-node hidden layer coupled with ReLu activation functions expect for the last layer.
This Q approximator is trained for 5000 episodes, each with a horizon of 100. Evaluation mirrored the method described previously, with 50 repetitions conducted to ascertain the average performance metrics.

\begin{figure}[t]
    \centering
    \includegraphics[width=0.8\linewidth]{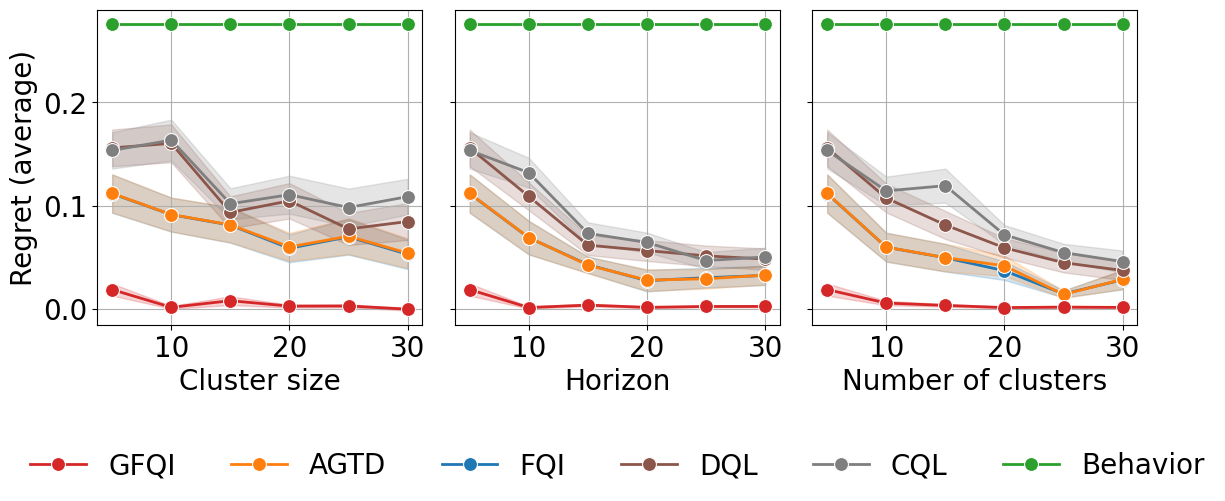}
    \caption{Change in regret of the average reward with varying cluster size, time horizon or number of clusters. The band represents the standard error calculated with respect to the random seed after running experiments 50 times.}
    \label{fig:sim_regret_average_team_size}
\end{figure}

\noindent \textbf{Results.}
The convergence patterns and regret metrics on the average reward with are depicted in the Figure \ref{fig:sim_regret_average_team_size}. 
Similar to the results in Section \ref{sec:numirical}, the GFQI estimator achieved the fastest convergence in regret and out performs the neural network based estimator (CQL and DDQN) in most of the cases.

\end{document}